\documentclass[lettersize,journal,romanappendices]{IEEEtran}
\usepackage{amsmath,amsfonts}
\usepackage{algorithmic}
\usepackage{algorithm}
\usepackage{array}
\usepackage[caption=false,font=normalsize,labelfont=sf,textfont=sf]{subfig}
\usepackage{textcomp}
\usepackage{stfloats}
\usepackage{url}
\usepackage{verbatim}
\usepackage{graphicx}
\usepackage{cite}
\usepackage{makecell}
\usepackage{booktabs}
\usepackage{multirow}
\usepackage{tabularx}
\usepackage{color}
\usepackage{pifont}
\usepackage{colortbl}
\usepackage{orcidlink}
\usepackage{fontawesome5}

\definecolor{improvebg}{RGB}{237,246,237}   
\definecolor{improvefg}{RGB}{56,142,60}

\begin{document}

\title{\texorpdfstring{%
  \begingroup
  \setbox0=\hbox{\includegraphics[height=2.35em]{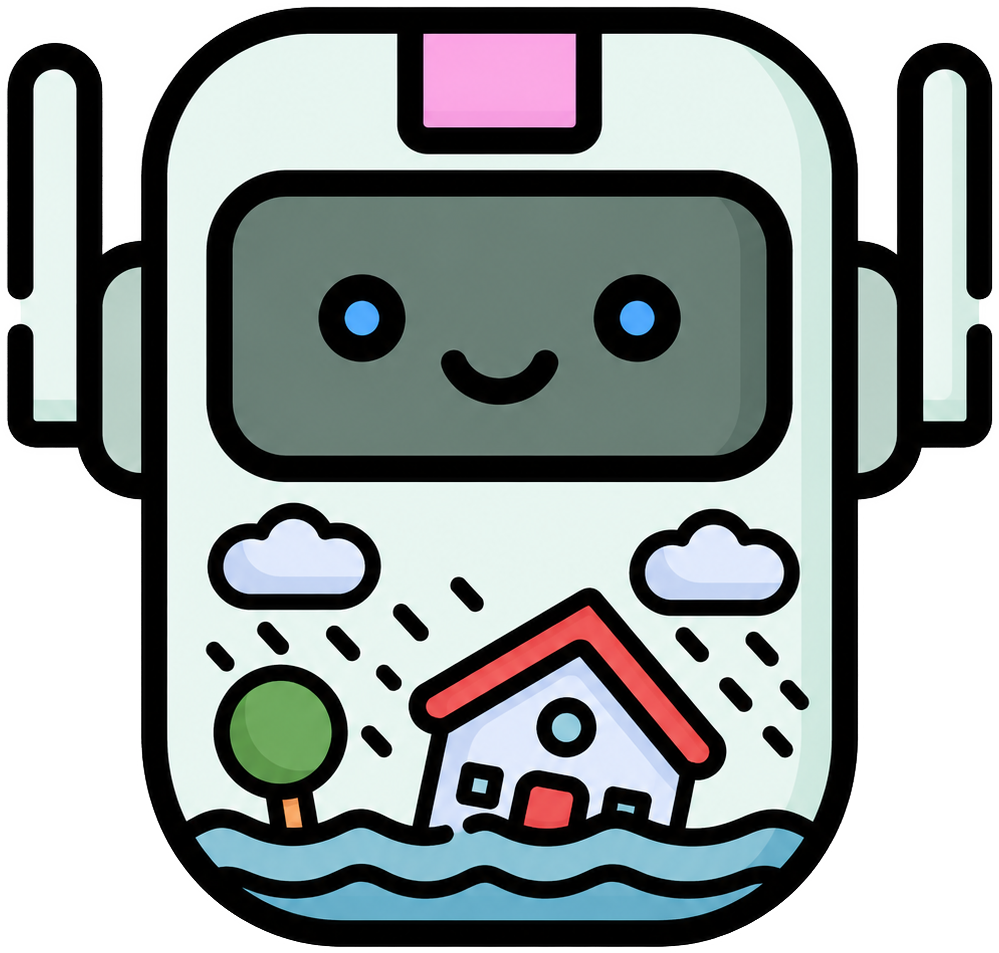}}%
  \begin{tabular}{@{\hspace{0.3em}}c@{\hspace{0.15em}}c@{}}%
    \raisebox{-0.85em}[0pt][0pt]{\usebox0} &
    \parbox[c]{\dimexpr\linewidth-\wd0-0.55em\relax}{%
      \centering\Huge
      DisasterAgent: An Evidence-Driven Agent for Verifiable Disaster Report Generation From Remote Sensing Imagery}%
  \end{tabular}%
  \endgroup
}{DisasterAgent: An Evidence-Driven Agent for Verifiable Disaster Report Generation From Remote Sensing Imagery}}

\author{
  Zhijian~Li, Jiawei~Song, Chao~Ren, Yidan~Zhang, Peijin~Wang, Xian~Sun,~\IEEEmembership{Senior Member,~IEEE,}\\
and Kun~Fu,~\IEEEmembership{Senior Member,~IEEE}%
  \thanks{This work was supported by the National Natural Science Foundation of China under Grant 62425115. (Corresponding author: Kun Fu)

  Zhijian Li, Jiawei Song, Yidan Zhang, Peijin Wang, Xian Sun and Kun Fu are with the Aerospace Information Research Institute, Chinese Academy of Sciences, Beijing 100190, China; the School of Electronic, Electrical and Communication Engineering, University of Chinese Academy of Sciences, Beijing 100190, China; the University of Chinese Academy of Sciences, Beijing 100190, China; and the National Key Laboratory of Target Cognition and Application Technology (TCAT), Aerospace Information Research Institute, Chinese Academy of Sciences, Beijing 100190, China (e-mail: lizhijian25@mails.ucas.ac.cn, songjiawei25@mails.ucas.ac.cn, zhangyidan19@mails.ucas.ac.cn, wangpeijin17@mails.ucas.ac.cn, sunxian@aircas.ac.cn, fukun@aircas.ac.cn  ).

  Chao Ren is with the Aerospace Information Research Institute, Chinese Academy of Sciences, Beijing 100190, China; and the National Key Laboratory of Target Cognition and Application Technology (TCAT), Aerospace Information Research Institute, Chinese Academy of Sciences, Beijing 100190, China(e-mail: renc0003@e.ntu.edu.sg).
  }%
  \\[0.55em]
  {\normalsize\hypersetup{pdfborder={0 0 0}}%
  \textcolor{gray}{\faIcon{globe}}~\textcolor{blue}{Project Page:~\url{https://zhijianli2003.github.io/DisasterAgent}}}%
}

\IEEEaftertitletext{\vspace{-25pt}}
\maketitle

\begin{abstract}

Remote sensing disaster assessment is moving from simple damage classification toward comprehensive report-level analysis. In this setting, responders need not only damage descriptions but also verifiable evidence to guide response and recovery. Recent vision-language models (VLMs) and tool-augmented agents offer strong multimodal reasoning and perception, but they struggle to provide the evidence required for high-stakes verification. To address these limitations, we propose DisasterAgent, an evidence-driven agent that uses a Perception-Augmentation-Generation workflow to generate verifiable disaster reports from bitemporal optical imagery. DisasterAgent first invokes a specialized vision expert toolkit in the perception phase to build a Spatiotemporal Perception Graph (STPG), serving as a structured memory to store and organize disaster-induced changes across space and time. Evidence-based Augmentation (EA) then extracts evidence from the STPG, attaching inline citations that form a closed evidence loop from each statement to the supporting visual evidence. Extensive experiments on the DisasterM3 and RSCC benchmarks demonstrate that DisasterAgent consistently outperforms general-domain VLMs, remote sensing-specialized models, and existing remote sensing agents. Notably, compared with the original model, it improves the average score for disaster report generation on DisasterM3 by 13.5\% and ST5-SCS on RSCC by 15.34\%, providing a more accurate and verifiable solution for disaster emergency response. The code will be available at \url{https://github.com/ZhijianLi2003/DisasterAgent}.

\end{abstract}


\section{Introduction}

\begin{figure}[!t]
\centering
\includegraphics[width=\columnwidth]{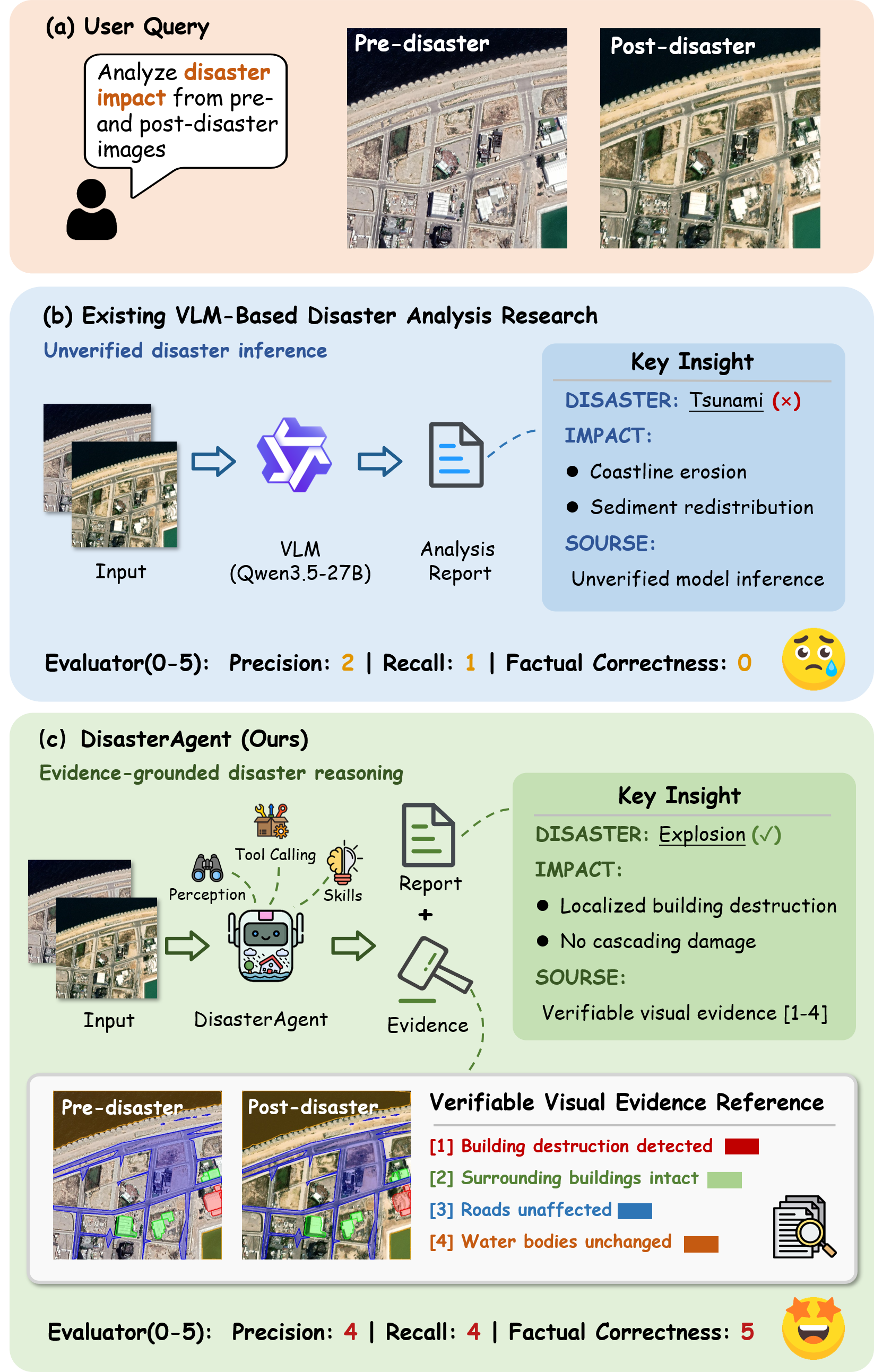}
\caption{Comparison between standard VLM inference and the evidence-grounded reasoning of DisasterAgent. (a) A user query for a disaster impact caption from a bitemporal optical image pair. (b) An existing VLM generates an unverified report and misidentifies the disaster type. (c) DisasterAgent invokes perception tools and attaches inline citations to pixel-level visual evidence, mitigating hallucinations and producing more accurate captions.}
\label{fig1_motivation}
\vspace{-8pt}
\end{figure}

Remote sensing imagery has become an indispensable data source for rapid disaster assessment, owing to its extensive spatial coverage, frequent revisit capabilities, and timely capture of large-scale post-disaster impacts \cite{kaku2019satellite, chen2025bright}. In particular, bitemporal remote sensing images provide crucial visual evidence for identifying disaster-induced changes by comparing pre- and post-event observations. Recent studies on remote sensing disaster assessment have achieved substantial progress in tasks such as damaged building detection, semantic change detection, and damage severity classification \cite{fujita2017damage, gupta2019creating, shen2021bdanet}. However, these methods typically formulate disaster assessment as a pixel-level or object-level perception problem, generating low-level structured outputs such as segmentation masks, bounding boxes, or predefined categorical labels \cite{xiao2023cross}.

Despite their effectiveness in quantitative damage mapping \cite{zheng2021building, chen2024changemamba}, such perception-oriented methods remain insufficient for high-level decision-making in real-world emergency response. In practical disaster management \cite{xu2025implementing}, analysts and responders require more than just localized damage metrics; they often require comprehensive, human-readable reports that summarize the disaster type, spatial distribution of damage, affected infrastructure, environmental changes, and actionable recovery priorities \cite{xu2025large, ndigande2026survey}. This requirement is especially critical in complex post-disaster scenarios where multiple land-cover categories, infrastructure systems, and secondary impacts intertwine. Consequently, remote sensing disaster assessment is moving from isolated visual perception toward high-level semantic reasoning and comprehensive report generation.

Recent advances in vision-language models (VLMs) provide a natural way to support this transition. By integrating visual perception with natural language reasoning, VLMs enable flexible interaction with remote sensing imagery, supporting tasks such as visual question answering, change captioning, and disaster report generation \cite{zhang2024earthgpt, kuckreja2024geochat, irvin2024teochat}. The emergence of specialized vision-language datasets \cite{rahnemoonfar2021floodnet, chen2026rscc, wang2026disasterm3} has further accelerated this trend by extending disaster assessment from conventional classification to language-based interpretation. Compared with traditional task-specific models, VLMs are highly flexible in handling open-ended queries and can generate textual descriptions that are more accessible to non-expert responders.

However, directly applying VLMs to remote sensing disaster report generation poses significant reliability risks. Most VLMs operate under an end-to-end generation paradigm, mapping visual and textual inputs directly to textual outputs without explicitly exposing the underlying visual evidence or reasoning process \cite{kuckreja2024geochat, irvin2024teochat}.  This black-box nature makes it exceedingly difficult to verify whether a generated statement is genuinely supported by the imagery. Consequently, VLMs may produce hallucinated or weakly grounded descriptions, especially in complex post-disaster scenes where subtle spatial changes, small damaged objects, and visually similar disaster patterns can easily trigger fabricated descriptions. While fine-tuning VLMs on disaster-specific datasets \cite{chen2026rscc, wang2026disasterm3} can partially mitigate these issues, it remains a resource-intensive process that fails to resolve the fundamental lack of explicit evidence grounding. In high-stakes emergency response, such unverifiable generation can limit their practical reliability.

To improve grounding and verifiability, recent studies have explored tool-augmented agents \cite{talemi2026agentic, kao2026towards} for remote sensing analysis. These agents typically employ large language models (LLMs) or VLMs as cognitive planners to invoke external perception tools, such as segmentation or damage assessment models, thereby obtaining more fine-grained information than pure end-to-end VLMs. However, when applied to high-stakes disaster assessment, existing remote sensing agents still face two key limitations. On the one hand, their capabilities are largely confined to elementary analytical tasks, such as single-image building counting or damage classification \cite{xu2024rs, feng2026earth}, rather than the comprehensive spatiotemporal reasoning required for disaster report generation. On the other hand, their tool outputs are often organized as discrete intermediate results without a unified spatiotemporal structure. This disjointed execution paradigm fails to organize scattered visual information into a coherent reasoning chain, rendering the provenance of generated statements untraceable and making verification difficult.

To address the aforementioned challenges, we propose DisasterAgent, an evidence-driven agent for verifiable disaster report generation from bitemporal remote sensing imagery. As illustrated in Fig.~\ref{fig1_motivation}, DisasterAgent organizes visual perception outputs into structured evidence and uses them to guide verifiable report generation, instead of relying on unconstrained end-to-end generation or fragmented tool-calling results. By anchoring generated statements to traceable spatiotemporal and historical evidence, DisasterAgent improves the factual reliability and verifiability of disaster reports.

The main contributions of this paper are summarized as follows:
\begin{itemize}
    \item We propose DisasterAgent, an evidence-driven agent that uses a Perception-Augmentation-Generation workflow to perform comprehensive disaster assessment. The architecture enables pre-trained VLMs to serve as verifiable disaster report generators without fine-tuning for specific tasks, notably improving the average report generation score on the DisasterM3 benchmark by 13.5\% and ST5-SCS on the RSCC benchmark by 15.34\%.
     
    \item To accurately perceive changes caused by disasters, we construct a specialized vision expert toolkit tailored for disaster assessment and design the Spatiotemporal Perception Graph (STPG). Serving as a structured memory, the STPG systematically organizes fragmented tool outputs into coherent spatial relationships and temporal evolution edges across phases.

    \item To enhance the verifiability of the generated reports, we propose a multi-source Evidence-based Augmentation (EA) mechanism, built upon the STPG, that integrates semantic change claims, pixel-level spatial grounding, and historical experience. Inline citations form a closed evidence loop from each statement to the supporting visual evidence. By anchoring factual claims to this traceable evidence, the mechanism mitigates VLM hallucinations and better supports reliable emergency response.
    
\end{itemize}

\section{Related Work}
\subsection{Remote Sensing Disaster Assessment}

Traditional remote sensing disaster assessment has predominantly been formulated as dense prediction or object detection tasks. The release of large-scale annotated datasets, such as ISBDA \cite{fujita2017damage}, xBD \cite{gupta2019creating}, and EBD \cite{wang2025constructing}, has driven the rapid development of deep learning methodologies in this domain. Researchers have extensively explored various architectures, including convolutional neural networks (CNNs) \cite{zheng2021building, shen2021bdanet}, vision transformers (ViTs) \cite{xiao2023cross}, and state space models (SSMs) \cite{chen2024changemamba}, to optimize bitemporal feature extraction and semantic change detection. While these methods achieve high precision in localizing damage and grading severity, they are inherently designed to output low-level visual formats (e.g., pixel-level masks). 

Furthermore, conventional models are typically tailored for specific disaster types \cite{zheng2021building} or isolated tasks \cite{rahnemoonfar2021floodnet}. However, real-world disasters often trigger cascading effects that simultaneously impact diverse geographical targets, making it difficult for disjointed, single-task models to synthesize a holistic view of the disaster scene. Consequently, there is a pressing need for a unified framework capable of integrating heterogeneous perception results \cite{gao2024cost, gao2025vision} to perform high-level semantic reasoning. Our work builds upon the robust perception capabilities of these specialized vision models but elevates the output to comprehensive, human-readable disaster reports.

\subsection{Vision-Language Models for Remote Sensing Disaster Understanding}

To support multimodal interpretation in remote sensing, several specialized datasets have been curated. FloodNet \cite{rahnemoonfar2021floodnet} provides a Visual Question Answering (VQA) benchmark for post-hurricane damage assessment. RSCC \cite{chen2026rscc} enables large-scale remote sensing change captioning. DisasterM3 \cite{wang2026disasterm3} further introduces a comprehensive benchmark for multi-task disaster assessment and long-form report generation.

Concurrently, foundational general-domain VLMs (e.g., LLaVA \cite{li2024llava}, Qwen-VL \cite{bai2025qwen3}, and InternVL \cite{wang2025internvl3}) have inspired the development of specialized remote sensing VLMs. EarthGPT \cite{zhang2024earthgpt} processes diverse multi-sensor data, GeoChat \cite{kuckreja2024geochat} introduces region-level grounded dialogue in static images, and TEOChat \cite{irvin2024teochat} extends conversational reasoning to bitemporal earth observation data for change detection.

To adapt these models to disaster scenarios, recent approaches \cite{jiang2025large, chen2026rscc, wang2026disasterm3} predominantly rely on instruction fine-tuning. However, this end-to-end generation paradigm has two main limitations. First, it operates as a black box, generating text without revealing pixel-level grounding, making the outputs difficult to verify. Second, it lacks extensibility, as adapting to novel disaster typologies requires resource-intensive data curation and retraining. In contrast, DisasterAgent employs a training-free, evidence-driven framework that decouples visual perception from language generation, ensuring that reports are explicitly grounded in verifiable evidence and easily extensible to new scenarios.

\subsection{Tool-Augmented Agents for Remote Sensing Analysis}

Tool-augmented agents have emerged as a promising solution to extend the capabilities of LLMs and VLMs by invoking external expert models \cite{talemi2026agentic}. By extracting instance-level or pixel-level grounding information, these agents extend their capabilities beyond pure text generation. For example, ThinkGeo \cite{shabbir2025thinkgeo} and OpenEarthAgent \cite{shabbir2026openearthagent} act as overarching planners that invoke specialized temporal VLMs (such as TEOChat) to handle multi-temporal analysis. Similarly, RS-Agent \cite{xu2024rs} and EarthAgent \cite{feng2026earth} specifically target disaster scenarios by routing queries to deep learning models like ChangeOS to conduct precise building damage assessment and severity classification. Other systems explore diverse execution mechanisms: Change-Agent \cite{liu2024change} enables interactive visual interpretation by dynamically calling proprietary segmentation tools; IC-EO \cite{lahouel2026ic} executes tool invocations by translating user queries into direct Python code; and CangLing-KnowFlow \cite{chen2025cangling} streamlines execution by retrieving and applying predefined workflows from an established knowledge base. However, these existing agents typically rely on linear tool-calling sequences, resulting in isolated intermediate outputs that are difficult to synthesize for complex disaster assessment. DisasterAgent distinguishes itself by introducing the STPG, which systematically organizes fragmented tool outputs into a unified, spatiotemporal-aware structure to guide coherent and verifiable report generation.

\section{Methodology}

\subsection{Overview of DisasterAgent}

\begin{figure*}[!t]
  \centering
  \includegraphics[width=\textwidth]{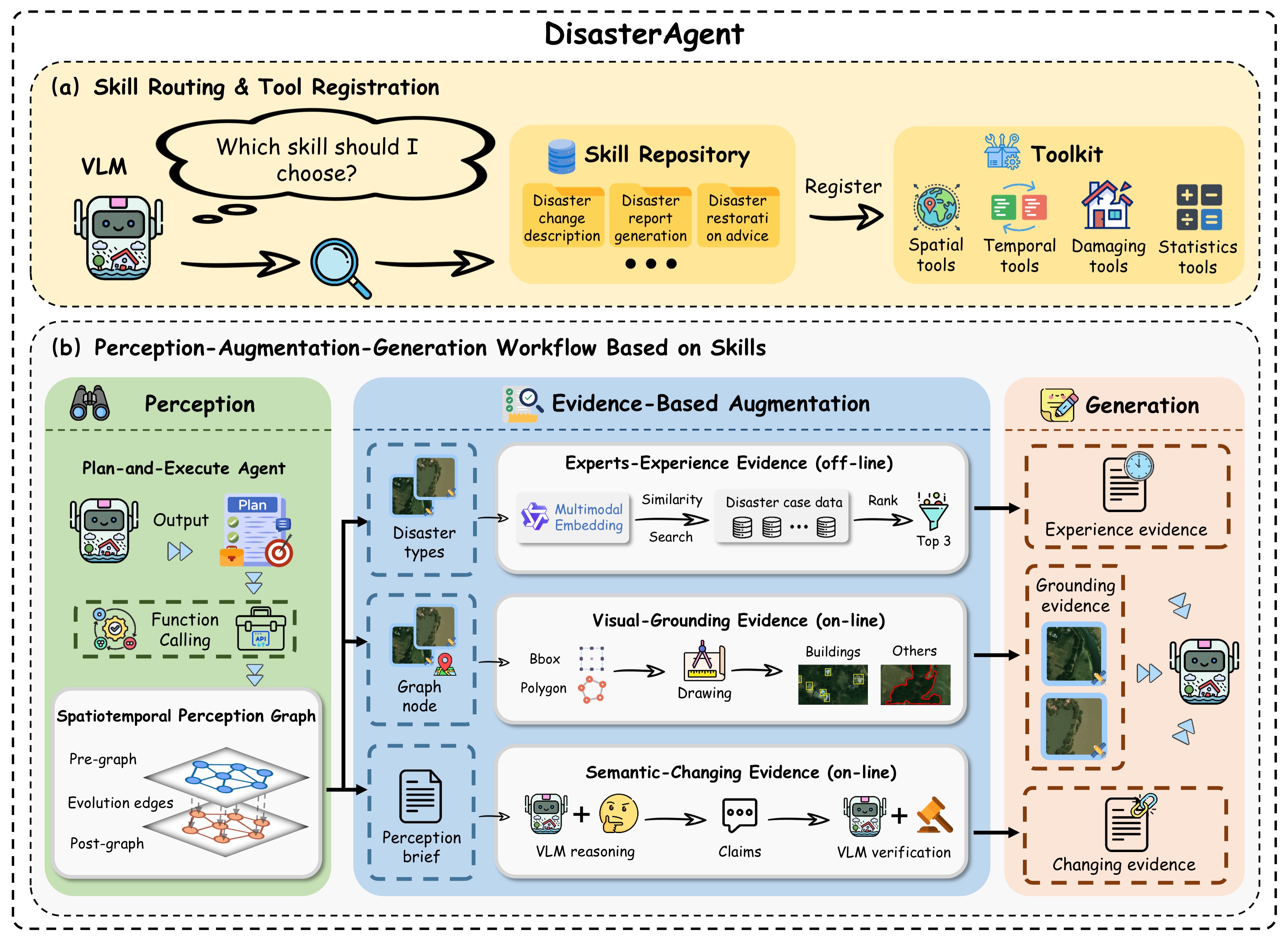}
  \caption{The overall architecture of DisasterAgent. The framework processes multimodal inputs through two primary stages. (1) Skill Routing: Driven by the VLM's interpretation of the query, the system dynamically routes to appropriate skills within a pre-defined repository and seamlessly registers the required functional tools. (2) Skill-Guided Workflow: Guided by the registered skills, the framework sequentially executes a three-step pipeline comprising Perception, Evidence-based Augmentation, and Generation.}
  \label{fig2_framework}
\end{figure*}

DisasterAgent is a skill-conditioned, evidence-driven agent that repurposes a single pre-trained VLM into a verifiable disaster report generator without the need for task-specific fine-tuning. As illustrated in Fig.~\ref{fig2_framework}, the system takes a bitemporal pair of remote sensing images (pre- and post-disaster) and a free-form natural language query as input. It processes these inputs through two primary stages. First, a Skill Routing phase dynamically selects the appropriate analytical skill and registers its required tools. Second, a Skill-Guided Workflow sequentially executes Spatiotemporal Change Perception, Evidence-based Augmentation, and Evidence-Fused Verifiable Generation. Ultimately, the system produces a comprehensive disaster report along with a list of citable references, grounding the generated statements in concrete visual and semantic evidence. The design of this three-step workflow is conceptually inspired by the Retrieval-Augmented Generation (RAG) paradigm \cite{lewis2020retrieval, edge2024local} prevalent in natural language processing (NLP). However, since remote sensing analysis fundamentally relies on precise visual grounding and spatiotemporal awareness rather than mere text retrieval, we tailor this concept to the unique demands of Earth Observation. By evolving the traditional RAG pipeline into a \textit{Perception-Augmentation-Generation} architecture, the framework establishes a structured, spatiotemporally aware reasoning process driven entirely by a central VLM. The entire workflow is orchestrated as a multi-stage process over a shared state:

\begin{enumerate}
    \item \textbf{Skill Routing:} The VLM inspects the input query and images to select the most appropriate skill from a Skill Repository (e.g., disaster change description, disaster report generation, or disaster restoration advice). This skill then registers a specialized toolkit comprising spatial, temporal, and damage assessment tools necessary for subsequent analysis.
    \item \textbf{Spatiotemporal Change Perception:} Guided by the selected skill, a Plan-and-Execute agent takes over, using function calling to systematically invoke the registered tools. It extracts information to instantiate a Spatiotemporal Perception Graph (STPG), which jointly captures pre-disaster structures, post-disaster structures, and their cross-phase evolution edges.
    \item \textbf{Evidence-based Augmentation:} The STPG (serialized as a perception brief and individual graph nodes) drives the extraction of three complementary evidence streams:

    \begin{itemize}
        \item \textit{Experts-Experience Evidence (EEE):} Retrieves the top-ranked historical disaster case data off-line via multimodal embedding similarity search, providing domain-specific priors and restoration guidance.
        \item \textit{Visual-Grounding Evidence (VGE):} Renders precise spatial annotations on-line, such as bounding boxes for buildings and polygons for other regions, directly onto the images.
        \item \textit{Semantic-Changing Evidence (SCE):} Reasons over the STPG on-line to generate atomic spatiotemporal claims. Each claim is independently verified by the VLM to obtain a confidence score, filtering out hallucinations and yielding robust changing evidence.
    \end{itemize}
    \item \textbf{Evidence-Fused Verifiable Generation:} Finally, the VLM fuses the experience evidence, grounding evidence, and changing evidence to generate the final response. The output includes the comprehensive disaster report with inline citations and a corresponding reference list that maps each citation back to the verified evidence, ensuring the entire generation process is fully verifiable.
\end{enumerate}

\subsection{Spatiotemporal Change Perception}
\subsubsection{Spatiotemporal Perception Graph}

\begin{figure}[!t]
  \centering
  \includegraphics[width=\linewidth]{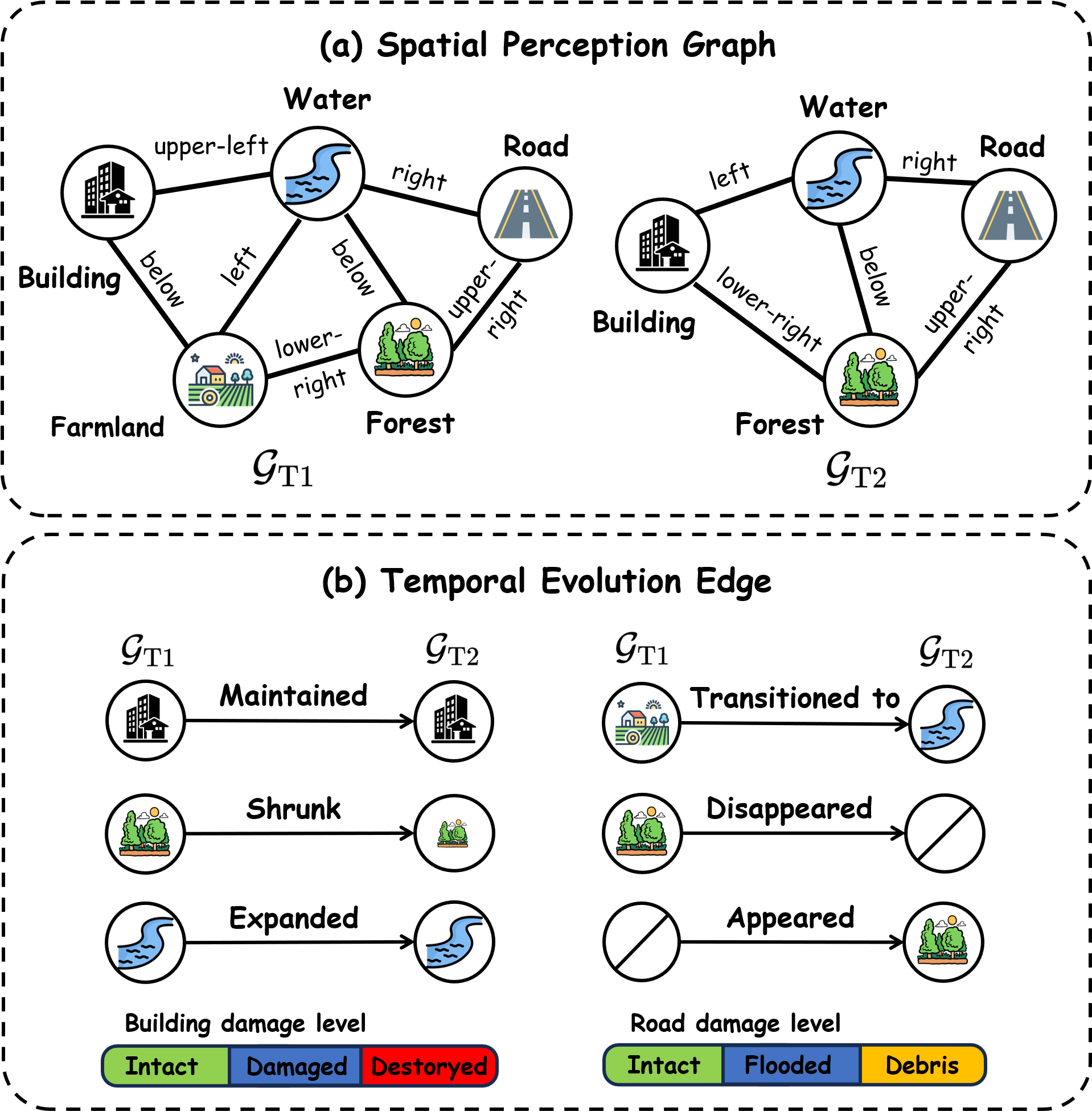} 
  \caption{Illustration of the STPG. (a) The Spatial Perception Graph constructs the spatial relationships between diverse geographical targets (e.g., buildings, roads, forests, farmland, and water) within a single phase. (b) The Temporal Evolution Edges establish the cross-phase mappings, explicitly capturing state transitions such as ``maintained'', ``shrunk'', ``expanded'', ``transitioned to'', ``disappeared'', and ``appeared'', while integrating domain-specific damage level assessments (e.g., building damage, road damage).}
  \label{fig3_stpg}
\end{figure}

To move beyond unstructured raw pixels while preserving spatially and temporally grounded evidence, we construct the STPG to represent the disaster scene. A key purpose of the STPG is to structurally store the outputs from various agent tool calls, facilitating the verification of low-level visual information during the reasoning process. As illustrated in Fig.~\ref{fig3_stpg}, the STPG systematically organizes the visual elements into single-phase perception graphs (representing the pre- and post-disaster spatial structures) and cross-phase evolution edges (capturing the temporal changes). Formally, the STPG is defined as:
\begin{equation}
\mathcal{G}_{\text{STPG}} = \big(\mathcal{G}_{\text{T1}},\, \mathcal{G}_{\text{T2}},\, \mathcal{E}_{\text{evo}}\big),
\end{equation}
where $\mathcal{G}_{\text{T1}}$ and $\mathcal{G}_{\text{T2}}$ denote the spatial perception graphs for the pre-disaster (Time 1) and post-disaster (Time 2) images, respectively, and $\mathcal{E}_{\text{evo}}$ is the set of temporal evolution edges connecting them. Notably, this formulation is highly extensible and can be naturally generalized to multi-temporal scenarios by chaining additional temporal graphs and evolution edges, e.g., $\big(\mathcal{G}_{\text{T1}}, \mathcal{E}_{\text{evo}}^{(1)}, \mathcal{G}_{\text{T2}}, \dots, \mathcal{E}_{\text{evo}}^{(n-1)}, \mathcal{G}_{\text{Tn}}\big)$.

\textbf{Spatial Perception Graph.} For a single phase (either pre- or post-disaster), the graph $\mathcal{G}_{*} = (\mathcal{V}_{*}, \mathcal{R}_{*})$ consists of region nodes and their spatial relation edges. Each node $v_i \in \mathcal{V}_{*}$ is defined as:
\begin{equation}
v_i = \big(\mathit{id}_i,\, \ell_i,\, b_i,\, m_i,\, \sigma_i,\, \boldsymbol{\phi}_i\big),
\end{equation}
where $\mathit{id}_i$ is a unique identifier of the region node, $\ell_i$ is the normalized semantic label (e.g., building, road, farmland, water), $b_i = (x,y,w,h)$ is the axis-aligned bounding box, $m_i$ represents the COCO-RLE-encoded segmentation mask, $\sigma_i \in [0,1]$ is the detection confidence score, and $\boldsymbol{\phi}_i$ is an extensible property dictionary for storing damage labels. We establish directed edges to connect each region to its $k$-nearest neighbors based on spatial proximity. In our implementation, we set $k=3$. The relation edge $r_{ij} \in \mathcal{R}_{*}$ is formulated as:
\begin{equation}
r_{ij} = \big(v_i,\, \rho(v_i, v_j),\, v_j\big), \quad \rho = \mathcal{R}_{\theta}\big(\angle(p_j - p_i)\big),
\end{equation}
where $p_i$ and $p_j$ denote the geometric centers of regions $v_i$ and $v_j$ respectively, and $\mathcal{R}_{\theta}(\cdot)$ quantizes the angle $\angle(p_j - p_i)$ between them into 8 discrete spatial directions (e.g., right, upper-right,
above).

\textbf{Temporal Evolution Edge.} To establish accurate temporal mappings across phases, we employ a mask-based region Intersection over Union (IoU) operator:
\begin{equation}
\operatorname{IoU}(v_i, v_j) = \frac{|m_i \cap m_j|}{|m_i \cup m_j|}.
\end{equation}
Similarly, the area of the mask region corresponding to node $v_i$ is $|v_i| = |m_i|$. For each region $v_i \in \mathcal{V}_{\text{T1}}$ in the pre-disaster graph $\mathcal{G}_{\text{T1}}$, we determine its optimal matching region $v^{\star}_i$ in the post-disaster graph $\mathcal{G}_{\text{T2}}$:
\begin{equation}
v^{\star}_i = \arg\max_{v_j \in \mathcal{V}_{\text{T2}}} \operatorname{IoU}\big(v_i, v_j\big), \quad u_i = \operatorname{IoU}\big(v_i, v^{\star}_i\big).
\end{equation}
We define the area-change ratio as $\alpha_i = |v^{\star}_i| / |v_i|$. The evolution type $\eta_i$ for the edge connecting $v_i$ and $v^{\star}_i$ is assigned according to the following piecewise rule:
\begin{equation}
\eta_i =
\begin{cases}
\textit{Disappeared}, & u_i < \tau_{\text{IoU}},\\[2pt]
\textit{Transitioned to}, & (u_i \ge \tau_{\text{IoU}}) \wedge (\ell_i \neq \ell^{\star}_i),\\[2pt]
\textit{Expanded}, & (u_i \ge \tau_{\text{IoU}}) \wedge (\ell_i = \ell^{\star}_i) \wedge (\alpha_i > \tau_{\uparrow}),\\[2pt]
\textit{Shrunk}, & (u_i \ge \tau_{\text{IoU}}) \wedge (\ell_i = \ell^{\star}_i) \wedge (\alpha_i < \tau_{\downarrow}),\\[2pt]
\textit{Maintained}, & \text{otherwise,}
\end{cases}
\end{equation}
where $\ell^{\star}_i$ denotes the semantic label of $v^{\star}_i$, and $\tau_{\text{IoU}}$, $\tau_{\uparrow}$, and $\tau_{\downarrow}$ are predefined thresholds (set to $\tau_{\text{IoU}}=0.3$, $\tau_{\uparrow}=1.5$, and $\tau_{\downarrow}=0.667$ by default). Regions in $\mathcal{G}_{\text{T2}}$ that fail to match any region in $\mathcal{G}_{\text{T1}}$ (i.e., $\max_i \operatorname{IoU}(v_i, v_j) < \tau_{\text{IoU}}$) are designated as \textit{Appeared}. The resulting cross-phase edge set $\mathcal{E}_{\text{evo}}$ thus explicitly encodes temporal dynamics such as existence, identity, and geometry transitions, providing foundational evidence for downstream reasoning.  

\subsubsection{Tool-augmented Visual Perception}

Building the aforementioned $\mathcal{G}_{\text{STPG}}$ requires fusing heterogeneous pixel-level vision models. To achieve this in a decoupled and extensible manner, we delegate the execution to a Plan-and-Execute agent. In this paradigm, the VLM acts purely as a \textit{planner}, while a sandbox Python runtime acts as the \textit{executor}. 

Given the task description and the skill configuration, the planner emits a structured JSON tool-call plan:
\begin{equation}
\pi = \big[(t_1,\mathbf{a}_1),\, (t_2,\mathbf{a}_2),\, \dots\big],
\end{equation}
where each step references a registered tool $t_k$ with arguments $\mathbf{a}_k$. A critical feature of this orchestration strategy is the cross-step data flow expressed via symbolic references (e.g., \texttt{"\$step\_3"}). This allows the STPG to be sequentially updated and threaded through successive tool calls without the VLM needing to handle raw intermediate data. The runtime then executes the plan deterministically, captures all return values and errors, and serializes both the plan and the execution log for later verification.

The registered toolkit is categorized into four functional families: 1) \textbf{Spatial tools} ingest open-vocabulary instance-level segmentation results to instantiate region nodes in $\mathcal{G}_{\text{T1}}$ and $\mathcal{G}_{\text{T2}}$; 2) \textbf{Temporal tools} compute mask-based IoU to construct the cross-phase evolution edges; 3) \textbf{Damage tools} extract specific disaster impacts (e.g., building damage, road submersion) and project these semantic masks onto the STPG via pixel-voting to update node properties $\boldsymbol{\phi}_i$; and 4) \textbf{Statistics tools} aggregate the graph content into compact numerical or textual summaries to facilitate downstream VQA-style reasoning.

The specific vision expert models underlying these tools are detailed in Table~\ref{tab:vision_experts}. To obtain more accurate pixel-level damage information, some of these tools are pre-trained on the DisasterM3-Instruct and xBD datasets. Because the VLM only plans and never executes the tools directly, tool injection is fully sandboxed. This design ensures that the agent can generalize to new skills or disaster queries simply by modifying the configuration prompt, without requiring any changes to the underlying tool implementations. The designs of the disaster type recognition and building damage assessment tools are provided in Appendix~\ref{app:tools}, as shown in Fig.~\ref{fig:app_tools}.

\begin{table*}[!t]
  \centering
  \caption{Vision Expert Tools for Spatiotemporal Change Perception}
  \label{tab:vision_experts}
  \renewcommand{\arraystretch}{1.15}
  \begin{tabular}{p{0.28\linewidth} p{0.64\linewidth}}
      \toprule
      \multicolumn{1}{c}{\textbf{Vision Expert Tool}} 
      & \multicolumn{1}{c}{\textbf{Function Description}} \\
      \midrule
      
      Disaster Type Recognition \cite{simeoni2025dinov3}
      & Built upon DINOv3, this tool identifies the disaster type from bitemporal remote sensing images and supports 10 representative disaster categories, including earthquakes, floods, hurricanes, volcanoes, explosions, and conflicts. \\
      \addlinespace
      
      Building Damage Analysis (4-class) \cite{zheng2021building}
      & Implemented with ChangeOS, this tool performs pixel-level building damage assessment based on the Joint Damage Scale, which integrates FEMA and HAZUS standards, and categorizes building conditions into four severity levels: no-damage, minor-damage, major-damage, and destroyed. \\
      \addlinespace
      
      Building Damage Analysis (3-class) \cite{chen2024changemamba}
      & Based on ChangeMamba, this tool provides pixel-level structural damage perception following FEMA-oriented damage definitions, classifying buildings into three levels: intact, damaged, and destroyed. \\
      \addlinespace
      
      Road Damage Analysis \cite{chen2024changemamba}
      & Powered by ChangeMamba, this tool extracts pixel-level road damage information from pre- and post-disaster imagery, distinguishing undamaged roads, flood-covered roads, and roads damaged by debris or rubble. \\
      \addlinespace
      
      General Concept Segmentation \cite{carion2025sam}
      & Integrated with SAM3, this tool enables open-vocabulary object detection and segmentation for general disaster-related concepts, such as farmland, forest, water bodies, and other scene elements, supporting zero-shot generalization across diverse disaster scenarios. \\
      
      \bottomrule
  \end{tabular}
\end{table*}

\subsection{Evidence-based Augmentation}

To ensure the generated disaster reports are grounded and verifiable, we design three complementary evidence streams (as depicted in the Evidence-based Augmentation block of Fig.~\ref{fig2_framework}). These evidence modules are operationally divided into on-line extraction and off-line retrieval mechanisms. On-line, the framework operates directly on the STPG to dynamically extract SCE and VGE. Conceptually, SCE encodes \textit{what changed and why we believe so} at a high semantic level, while VGE provides precise spatial localization, revealing \textit{where it changed at the pixel level}. Off-line, EEE retrieves domain-specific historical priors from a pre-constructed knowledge base, demonstrating \textit{how similar past events were described and remediated}. Together, they form a robust, multi-dimensional evidence base for the subsequent report generation process.

\subsubsection{On-line Semantic-Changing Evidence Reasoning}

\begin{figure*}[!t]
  \centering
  \includegraphics[width=\linewidth]{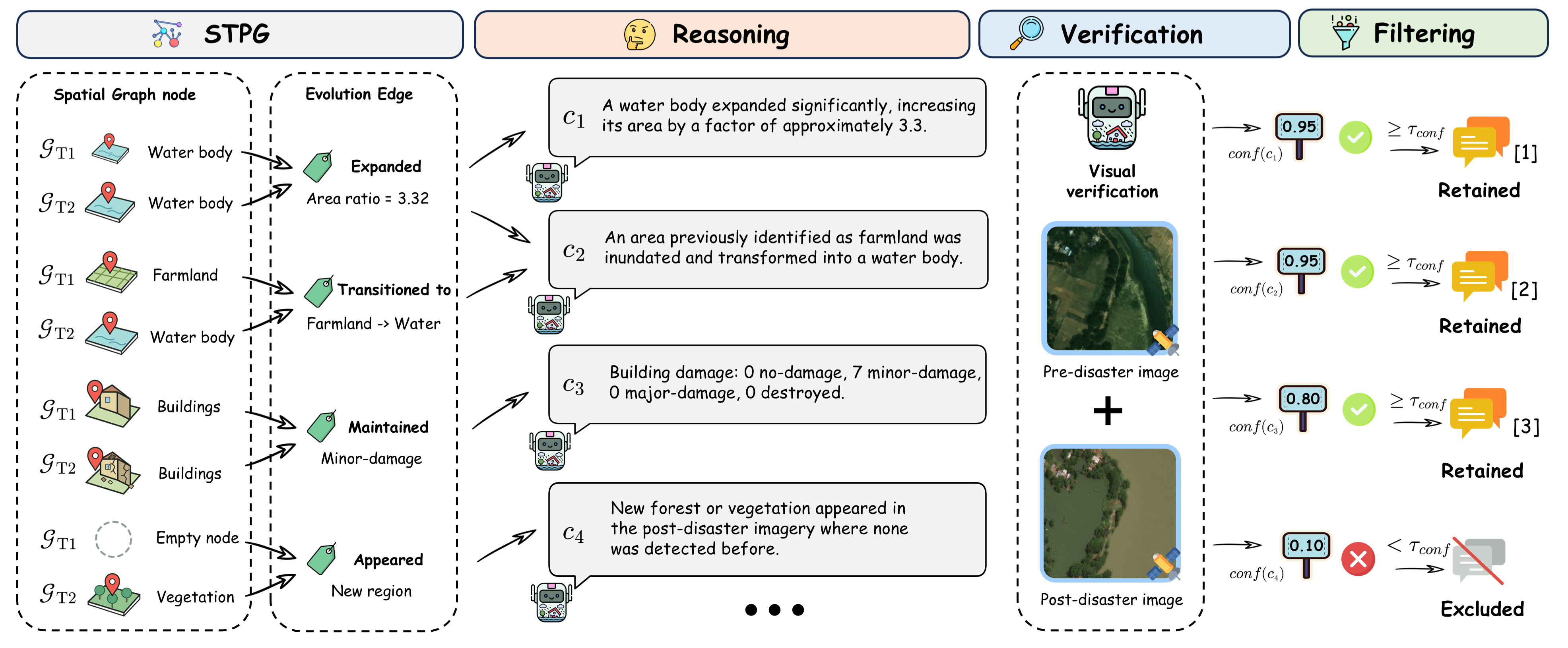} 
  \caption{Illustration of the SCE extraction pipeline. The VLM first reasons over the STPG to generate atomic factual claims. Each claim is then independently verified against the raw bitemporal images to obtain a confidence score. Finally, claims with scores above the threshold are filtered to form the robust SCE set.}
  \label{fig4_SCEget}
\end{figure*}

As illustrated in Fig.~\ref{fig4_SCEget}, SCE follows a \textit{reasoning $\to$ verification $\to$ filtering} pipeline to extract semantic change information from the STPG. This process is divided into three steps:

\begin{itemize}
    \item \textbf{Reasoning:} The central VLM reads the Perception Brief, which contains the disaster type, evolution edges, and damage labels, to emit a structured Evidence Chain. This chain consists of (i) a set of atomic, verifiable claims $\mathcal{C}=\{c_k\}$ in natural language (e.g., ``The river expanded and submerged adjacent farmland.''), and (ii) a list of evidence references for each claim that cites specific STPG entities (such as evolution edges or specific region nodes). Consequently, rather than generating free-form text, this reasoning step establishes a structured mapping where each claim $c_k \in \mathcal{C}$ is explicitly linked to its supporting STPG evidence references. This structured relationship is then serialized and stored in JSON format, facilitating subsequent verification.

    \item \textbf{Verification:} Although the claims derived from the reasoning step originate from the perception of visual expert tools and are relatively reliable, they may still contain inaccuracies due to upstream perception errors. To prevent the VLM from simply rubber-stamping these claims, an independent verification mechanism is introduced. This design is motivated by the observation that in complex remote sensing image interpretation tasks, VLMs generally exhibit stronger and more reliable performance as discriminators (i.e., verifying existing claims) than as free-form generators \cite{li2023evaluating,wu2026generate,li2025ddfav}. For every unverified claim $c_k$, a separate VLM call is executed using only the raw image pair $(I_{\text{T1}}, I_{\text{T2}})$ and the claim $c_k$ as input. Acting as a strict visual auditor, the model returns a confidence score. The returned scalar is written back into the claim node:
\begin{equation}
\text{conf}(c_k) \leftarrow \mathrm{VLM}_{\text{verify}}\big(c_k \mid I_{\text{T1}}, I_{\text{T2}}\big).
\end{equation}

    \item \textbf{Filtering:} Claims satisfying $\text{conf}(c_k) \geq \tau_{\text{conf}}$ (where $\tau_{\text{conf}} = 0.7$ by default) are retained to form the final semantic-changing evidence set $\mathcal{E}_{\text{SCE}}$. Formally, this set is defined as:
\begin{equation}
\mathcal{E}_{\text{SCE}} = \Big\{\, c_k \;\Big|\; \text{conf}(c_k) \ge \tau_{\text{conf}},\; c_k \in \mathcal{C} \,\Big\}.
\end{equation}
Each surviving claim then serves as a citable reference during the final report generation.
\end{itemize}

This design transforms a monolithic black-box VLM inference into a \textit{self-auditing} reasoning loop. Only the claims that simultaneously (i) align with the structured STPG and (ii) pass the independent VLM evaluation with a confidence score above the threshold are permitted to enter the final generation stage.

\subsubsection{On-line Visual-Grounding Evidence Extraction} 

While SCE supplies global, semantic-level evidence, VGE provides fine-grained, pixel-level evidence by directly rendering the STPG content onto the original images. Specifically, for every region in the spatial perception graphs $\mathcal{G}_{\text{T1}}$ and $\mathcal{G}_{\text{T2}}$ that carries a damage label, we generate precise spatial annotations. For example, buildings are annotated with axis-aligned bounding boxes, while roads and other irregular regions are delineated with polygon contours. The colors of these annotations intuitively encode their respective damage severity (e.g., intact, damaged, or destroyed). 

The resulting annotated images, alongside an automatically generated color legend, are subsequently integrated into the VLM's multimodal context. Formally, the visual-grounding evidence set $\mathcal{E}_{\text{VGE}}$ is defined as the union of the visually rendered region nodes and their corresponding textual legend $L$:
\begin{equation}
\mathcal{E}_{\text{VGE}} = \Big\{\, \operatorname{Render}(v_i) \;\Big|\; v_i \in \mathcal{V}_{\text{T1}} \cup \mathcal{V}_{\text{T2}} \,\Big\} \cup \{L\},
\end{equation}
where $\operatorname{Render}(\cdot)$ denotes the operation of drawing the bounding box $b_i$ or polygon mask $m_i$ of region $v_i$ onto the image using a color mapped from its damage property $\boldsymbol{\phi}_i$, and $L$ maps each color to its specific semantic meaning (e.g., ``Red: Destroyed Building''). This explicit visual grounding mechanism enables the model to directly attend to the structured perception outcomes anchored to specific instances. Consequently, VGE functions as a highly precise spatial evidence stream, which is particularly crucial for substantiating fine-grained damage quantities and spatial distribution patterns within the generated disaster report.

\subsubsection{Off-line Experts-Experience Evidence Retrieval}

Inspired by the practice of emergency-response analysts who consult historical casebooks when assessing a new event, this module supplements the agent's generation process with external domain expertise. By retrieving historical disaster cases relevant to the current visual findings, the system is provided with essential remote sensing disaster assessment priors that VLMs may not have encountered during their general pre-training phase. Furthermore, compared to task-specific VLM fine-tuning, this retrieval-based approach offers superior extensibility, as it allows the knowledge base to be updated dynamically without retraining the underlying model.

We construct the source corpus using the training data from the DisasterM3-Instruct dataset. The initial corpus is denoted as:
\begin{equation}
\mathcal{S}=\Big\{\big(I_{\text{T1}}^{(j)}, I_{\text{T2}}^{(j)}, y_j, D_j, R_j\big)\Big\}_{j=1}^{N},
\end{equation}
where $N$ is the total number of training cases, $I_{\text{T1}}^{(j)}$ and $I_{\text{T2}}^{(j)}$ are the pre- and post-disaster images of the $j$-th case, $y_j$ is its disaster-type label, and $D_j, R_j$ represent the corresponding expert-verified disaster description and restoration advice, respectively. To ensure efficient retrieval and maintain diversity, we first apply Term Frequency-Inverse Document Frequency (TF-IDF) \cite{sparck1972statistical} vectorization on the concatenated text $D_j \oplus R_j$ for each case in $\mathcal{S}$. Then, for each unique disaster type $y$, we perform $K$-means clustering on the corresponding sub-corpus $\mathcal{S}_y$ to obtain $M=50$ cluster centroids:

\begin{equation}
\min_{\{\boldsymbol{\mu}_k\}_{k=1}^M} \sum_{j \in \mathcal{S}_y} \min_{1 \le k \le M} \big\lVert \operatorname{TF-IDF}(D_j \oplus R_j) - \boldsymbol{\mu}_k \big\rVert^2.
\end{equation}
By sampling the centroid-nearest case from each cluster, we obtain a lexically and semantically diverse knowledge base $\mathcal{B}$ comprising 50 representative historical cases per disaster type. In this retrieval framework, the multimodal inputs $(I_{\text{T1}}^{(j)}, I_{\text{T2}}^{(j)}, y_j)$ act as the query keys, while the textual descriptions $(D_j, R_j)$ serve as the corresponding values to be retrieved.

We employ the Qwen3-VL-Embedding \cite{Li2026qwen3vlembedding} model, denoted as $E_{\text{mm}}(\cdot)$, to extract representations for the query keys. Conditioned on a specific retrieval instruction $\mathcal{I}^{*}$, the off-line embedding $\mathbf{e}_{j}$ for each historical case in $\mathcal{B}$ is calculated as:
\begin{equation}
\mathbf{e}_{j} = E_{\text{mm}}\!\left(I_{\text{T1}}^{(j)} \oplus I_{\text{T2}}^{(j)} \oplus y_j \;\big|\; \mathcal{I}^{*}\right),
\end{equation}
where $\oplus$ denotes multimodal sequence concatenation. 

During inference, the agent dynamically identifies the disaster type $\hat{y}$ via the previously constructed STPG. The query embedding $\mathbf{q}$ is generated on-line following the same embedding procedure:
\begin{equation}
\mathbf{q} = E_{\text{mm}}\!\left(I_{\text{T1}} \oplus I_{\text{T2}} \oplus \hat{y} \;\big|\; \mathcal{I}^{*}\right).
\end{equation}
The cosine similarity score $s_{j}$ between the query and each historical case in the knowledge base is computed as:
\begin{equation}
s_{j} = \frac{\mathbf{q} \cdot \mathbf{e}_{j}}{\lVert\mathbf{q}\rVert \lVert\mathbf{e}_{j}\rVert}, \quad \forall j \in \mathcal{B}.
\end{equation}
Finally, the evidence set $\mathcal{E}_{\text{EEE}}$ is constructed by retrieving the top-$K$ most relevant historical text values based on $s_{j}$:
\begin{equation}
\mathcal{E}_{\text{EEE}} = \Big\{\,(D_{j},\,R_{j}) \;\Big|\; \operatorname{rank}\!\big(s_{j}\big) \le K,\; j \in \mathcal{B}\,\Big\},
\end{equation}
where $\operatorname{rank}(\cdot)$ returns the descending rank of the similarity score in the candidate pool. By default, $K$ is set to 3. The retrieved descriptions and advice are subsequently split and injected into the multi-stage generation prompt, ensuring that historical descriptions inform factual disaster captioning while historical advice provides validated restoration recommendations.

\subsection{Evidence-Fused Verifiable Generation} 

\begin{figure}[htbp]
  \centering
  \includegraphics[width=\linewidth]{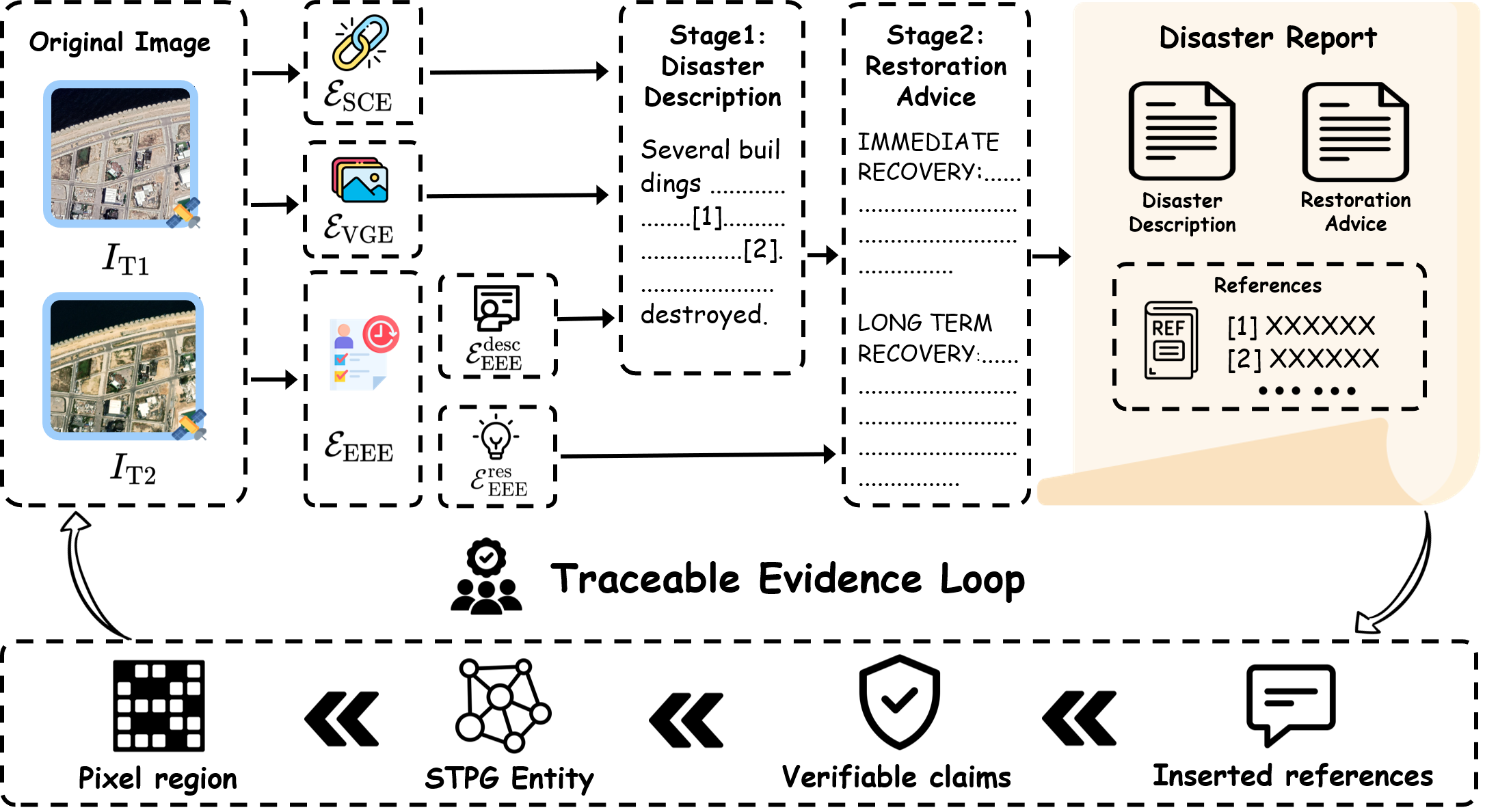} 
  \caption{Illustration of the Evidence-Fused Verifiable Generation. The framework employs a cascaded two-stage generation process: Stage 1 generates the disaster description grounded in all three evidence streams ($\mathcal{E}_{\text{SCE}}$, $\mathcal{E}_{\text{VGE}}$, $\mathcal{E}_{\text{EEE}}^{\text{desc}}$), while Stage 2 provides restoration advice conditioned exclusively on the generated description and the retrieved historical advice ($\mathcal{E}_{\text{EEE}}^{\text{res}}$). Through explicit inline citations (e.g., \textit{[1]}, \textit{[2]}) mapped to a reference list, the system establishes a closed, traceable evidence loop connecting the final report back to the raw pixel regions.}
  \label{fig_generation}
\end{figure}

The final generation stage fuses the three evidence streams ($\mathcal{E}_{\text{SCE}}$, $\mathcal{E}_{\text{VGE}}$, $\mathcal{E}_{\text{EEE}}$) with the raw image pair $(I_{\text{T1}}, I_{\text{T2}})$ to produce a comprehensive and structured disaster report. As illustrated in Fig.~\ref{fig_generation}, to ensure the resulting report is highly verifiable, we introduce two key design mechanisms: a cascaded two-stage generation architecture and an explicit per-claim citation system.

\textbf{Cascaded Two-Stage Generation.} Rather than generating the entire disaster report in a single, monolithic VLM call, we decompose the generation process into two cascaded stages. This design mirrors the cognitive process of human responders: first describing the factual impacts, then prescribing subsequent recovery actions:
\begin{equation}
\label{eq:desc}
R_{\text{desc}} = \mathrm{VLM}\!\big(I_{\text{T1}}, I_{\text{T2}},\, \mathcal{E}_{\text{SCE}},\, \mathcal{E}_{\text{VGE}},\, \mathcal{E}_{\text{EEE}}^{\text{desc}}\big),
\end{equation}
\begin{equation}
\label{eq:res}
R_{\text{res}} = \mathrm{VLM}\!\big(I_{\text{T1}}, I_{\text{T2}},\, R_{\text{desc}},\, \mathcal{E}_{\text{EEE}}^{\text{res}}\big),
\end{equation}
where $\mathcal{E}_{\text{EEE}}^{\text{desc}}$ and $\mathcal{E}_{\text{EEE}}^{\text{res}}$ represent the historical disaster descriptions and restoration advice fields retrieved from $\mathcal{E}_{\text{EEE}}$, respectively. 

In Stage 1 \eqref{eq:desc}, the VLM consumes all three evidence streams to ground the factual disaster description $R_{\text{desc}}$ in fine-grained visual details and verified semantic claims. Conversely, Stage 2 \eqref{eq:res} is deliberately evidence-restricted: the SCE and VGE streams are switched off. Instead, the newly generated $R_{\text{desc}}$ serves as the factual anchor, and only the historical advice field $\mathcal{E}_{\text{EEE}}^{\text{res}}$ is injected. This cascaded design yields significant benefits: (i) the restoration advice is strictly conditioned on a committed description supported by citations, effectively eliminating the risk of hallucinating fabricated damage during the recommendation phase; and (ii) the recommendations are guided by historically validated actions tailored to the current disaster type, rather than being inferred directly from raw pixels. This implements an explicit chain of thought at the system level.

\textbf{Traceable Evidence Loop.} In general-domain text generation, linking each statement to supporting evidence by citation is an effective way to make generated text verifiable \cite{menick2022teaching, gao2023enabling, slobodkin2024attribute}. To establish this provenance for disaster reports (shown at the bottom of Fig.~\ref{fig_generation}), the generation prompt is explicitly instructed to cite the indices of the supporting SCE claims when generating the report. Specifically, the model receives a formatted evidence block where claims are enumerated as \texttt{[i] <text> (confidence score: 0-1)}. The runtime automatically serializes the generated texts alongside a references list that maps every cited index back to the original claim text, confidence score, and underlying STPG entities. Consequently, human auditors can traverse this closed evidence loop in $O(1)$ complexity per claim to verify the factual accuracy of any statement within the report.

\section{Experiments}

To comprehensively evaluate the proposed DisasterAgent, we first introduce the benchmark datasets, evaluation metrics, and implementation details. We then present a quantitative performance comparison against existing VLMs and agent frameworks on two remote sensing disaster benchmarks. Finally, we conduct detailed ablation studies to validate our architectural designs and provide qualitative analyses to intuitively illustrate how DisasterAgent mitigates hallucinations and establishes a traceable evidence loop.

\subsection{Datasets}

We evaluate DisasterAgent on two complementary bitemporal remote sensing disaster benchmarks: RSCC~\cite{chen2026rscc} and DisasterM3~\cite{wang2026disasterm3}. Specifically, RSCC is utilized to examine fine-grained change captioning, while DisasterM3 provides a broader scope for comprehensive disaster assessment. A detailed comparison of these two datasets is provided in Table~\ref{tab:datasets}.

\begin{table}[!t]
  \centering
  \caption{Comparison of Benchmark Datasets Used for Evaluation}
  \label{tab:datasets}
  \renewcommand{\arraystretch}{1.3}
  \begin{tabular*}{\columnwidth}{@{\extracolsep{\fill}}ccc}
    \toprule
    & \textbf{RSCC~\cite{chen2026rscc}} & \textbf{DisasterM3~\cite{wang2026disasterm3}} \\
    \midrule
    \textbf{Tasks} & \makecell[c]{Change Captioning} & \makecell[c]{Disaster Type Recognition \\ Building Damage Counting \\ Disaster Region Estimation \\ Disaster Captioning \\ Restoration Advice} \\
    \midrule
    \textbf{\makecell[c]{Disaster \\ Types}} & \makecell[c]{Earthquake, Flood \\ Wildfire, Volcano \\ Storm, Tsunami} & \makecell[c]{Earthquake, Flood \\ Hurricane, Volcano \\ Explosion, Wildfire \\ Tornado, Tsunami} \\
    \midrule
    \textbf{Modality} & Optical & Optical \\
    \midrule
    \textbf{\makecell[c]{Test \\ Scale}} & \makecell[c]{988 image pairs} & \makecell[c]{2,512 image pairs} \\
    \bottomrule
  \end{tabular*}
\end{table}

\textbf{RSCC:} RSCC is a large scale change caption benchmark designed for understanding bitemporal images in disaster scenarios. Each sample consists of images taken before and after a disaster, along with a reference caption describing the corresponding scene changes. Compared with conventional caption datasets that focus on single image descriptions or generic land cover changes, RSCC emphasizes temporal transformations caused by disasters, including damaged buildings, inundated areas, vegetation loss, and infrastructure disruption. Following the official evaluation protocol, we evaluate DisasterAgent on the RSCC test benchmark, which contains 988 image pairs. This benchmark assesses whether the model can accurately capture and describe semantic changes from bitemporal observations.

\textbf{DisasterM3:} DisasterM3 is a comprehensive vision and language benchmark encompassing diverse hazards, sensors, and tasks for damage assessment. It contains 26,988 bitemporal satellite images and 123,010 instruction pairs collected from 36 historical events across five continents, covering ten common natural and anthropogenic disaster types. Unlike RSCC, DisasterM3 provides a broader scope for structured assessment and report generation. Its task taxonomy progresses from basic recognition and quantitative damage assessment to long form disaster description and restoration advice generation. In our experiments, we utilize the official DisasterM3 Bench set under the purely optical setting, focusing on the tasks most relevant to structured report generation.

\subsection{Evaluation Metrics}

We adopt task specific evaluation metrics for the two benchmarks, reflecting the different objectives of change captioning and comprehensive disaster report generation.

\textbf{Metrics on RSCC:} For the RSCC benchmark, we evaluate the generated change captions by comparing them with the reference captions using both lexical-overlap and semantic-similarity metrics. Specifically, we report ROUGE-L \cite{lin2004rouge} and METEOR \cite{banerjee2005meteor} as classical $n$-gram-based metrics. ROUGE-L measures the longest common subsequence between the generated and reference captions, while METEOR considers unigram matching with additional alignment and synonym-aware matching mechanisms. Since disaster reports can be semantically correct even when their wording differs from the reference caption, we also report ST5-SCS \cite{chen2026rscc}, an embedding-based contextual similarity metric. Following the RSCC benchmark, ST5-SCS encodes the generated and reference captions with Sentence-T5 \cite{ni2022sentence} and computes their sharpened cosine similarity. The formula is defined as:
\begin{equation}
\text{ST5-SCS} = \frac{1}{N} \sum_{i=1}^{N} \left| \frac{f_{\theta}(\hat{r}_i)^{\top} f_{\theta}(r_i)}{\|f_{\theta}(\hat{r}_i)\|_2 \|f_{\theta}(r_i)\|_2} \right|^3,
\end{equation}
where $N$ is the total number of evaluated samples, $\hat{r}_i$ and $r_i$ represent the generated caption and the reference caption for the $i$-th sample, respectively, and $f_{\theta}$ denotes the Sentence-T5-XXL text encoder. This metric is used to better capture semantic consistency in comprehensive disaster descriptions.

\begin{table}[!t]
  \centering
  \caption{Evaluation Dimensions for Comprehensive Report Generation}
  \label{tab:report_metrics}
  \renewcommand{\arraystretch}{1.3}
  \begin{tabular*}{\columnwidth}{@{\extracolsep{\fill}}llcc}
    \toprule
    \textbf{Task} & \textbf{Evaluation Dimension} & \textbf{Score} & \textbf{Abbr.} \\
    \midrule
    \multirow{3}{*}{\makecell[l]{Disaster \\ Captioning}} 
    & Damage Assessment Precision & $0\sim5$ & DAP \\
    & Damage Detail Recall & $0\sim5$ & DDR \\
    & Factual Correctness & $0\sim5$ & FC \\
    \midrule
    \multirow{3}{*}{\makecell[l]{Restoration \\ Advice}} 
    & Recovery Necessity & $0\sim5$ & RN \\
    & Action Priority Precision & $0\sim5$ & APP \\
    & Strategic Completeness & $0\sim5$ & SC \\
    \bottomrule
  \end{tabular*}
\end{table}

\textbf{Metrics on DisasterM3:} For the DisasterM3 benchmark, we report classification Accuracy (ACC) for Disaster Type Recognition (DTR). Unlike the original multiple choice setup, we formulate Building Damage Counting (BDC) and Disaster Region Estimation (DRE) as open ended numerical generation tasks to better reflect real world emergency response scenarios, where predefined candidate options are typically unavailable. We report the Mean Absolute Error (MAE) between the predicted and ground truth values for these tasks. For comprehensive report generation, we strictly adopt the official evaluation protocol and score the outputs with the official five point ($0 \sim 5$) rubric \cite{wang2026disasterm3}, evaluated by an LLM as a judge. The specific evaluation dimensions for Disaster Captioning and Restoration Advice are detailed in Table~\ref{tab:report_metrics}, with the overall score for each task computed as the average of its respective dimensions. Higher scores on this $0 \sim 5$ scale signify more accurate, comprehensive, and verifiable disaster assessment reports.

\subsection{Experimental Details}

DisasterAgent is implemented on top of the LangChain \cite{chase2022langchain} and LangGraph \cite{langgraph} frameworks. The entire four stage workflow, comprising Skill routing, Perception, Evidence-based Augmentation, and Generation, is compiled into a stateful directed graph using LangGraph. This architecture allows a shared, typed workflow state to flow seamlessly between nodes. Across all agentic stages, we reuse a single Qwen3.5-27B \cite{qwen_scope} as the central brain, accessed via the API provided by Alibaba Cloud's Bailian platform\footnote{\url{https://bailian.console.aliyun.com/}}. The unified VLM brain is wrapped as a standardized client that exposes text, multimodal, and JSON conversational interfaces. To ensure deterministic and reproducible outputs, we employ greedy decoding with the temperature set to $0.0$. Notably, no disaster specific fine-tuning is performed on this VLM.

Within the perception stage, the Plan-and-Execute agent generates structured JSON tool-call plans, which are executed by a local Python runtime through a unified tool interface.
The unified tool interface supports two execution modes: offline prediction cache and online model inference. For the offline prediction cache mode, the tool runtime pre-caches all possible prediction results for each tool call as JSON, which allows for fast and deterministic execution. For the online model inference mode, the tool runtime requests on-demand predictions from independently deployed FastAPI services over HTTP. The EEE retrieval service follows the same deployment as the online perception tool services.

\begin{table}[!t]
  \centering
  \caption{Performance Comparison on the RSCC Benchmark}
  \label{tab:RSCC}
  \renewcommand{\arraystretch}{1.3}
  \begin{tabularx}{\columnwidth}{Xccc}
      \toprule
      \multirow{3}{*}{Method} & \multicolumn{2}{c}{N-Gram} & \makecell{Contextual \\ Similarity} \\
      \cmidrule(lr){2-3} \cmidrule(lr){4-4}
       & \makecell{ROUGE-L \\ (\%)} & \makecell{METEOR \\ (\%)} & \makecell{ST5-SCS \\ (\%)} \\
      \midrule
      \multicolumn{4}{l}{\textcolor{blue}{$\bullet$} \textbf{General VLM}} \\
      Kimi K2.6 \cite{team2025kimi} & 11.14 & 10.98 & 52.29 \\
      GLM-4.5V \cite{hong2025glm} & 13.24 & 10.64 & 50.57 \\
      InternVL3.5-8B \cite{wang2025internvl3} & 12.05 & 10.84 & 47.98 \\
      Qwen3.5-122B-A10B \cite{qwen_scope} & 12.27 & 10.59 & 50.20 \\
      Qwen3.5-27B \cite{qwen_scope} & 11.68 & 11.00 & 51.84 \\
      Qwen3.5-9B \cite{qwen_scope} & 13.23 & 11.23 & 49.25 \\
      Qwen3.5-4B \cite{qwen_scope} & 12.82 & 12.06 & 50.59 \\
      \midrule
      \multicolumn{4}{l}{\textcolor{red}{$\bullet$} \textbf{Remote Sensing VLM}} \\
      CCExpert-7B \cite{wang2024ccexpert}        & 8.39 & 3.73 & 40.10 \\
      RSCCM-7B \cite{chen2026rscc} & \underline{14.99} & \textbf{16.05} & 58.52 \\
      \midrule
      \multicolumn{4}{l}{\textcolor[rgb]{0.50,0.20,0.65}{$\bullet$} \textbf{Remote Sensing Agent} \textnormal{\textit{(with the same expert toolkit as ours)}}} \\
      EarthAgent \cite{feng2026earth} & 15.31 & 13.50 & \underline{65.88} \\
      OpenEarthAgent \cite{shabbir2026openearthagent}        & \textbf{16.00} & 12.74 & 62.21 \\
      \midrule
      \multicolumn{4}{l}{\textcolor{green}{$\bullet$} \textbf{Ours}} \\
      DisasterAgent & \underline{15.76} & \underline{14.77} & \textbf{67.18} \\
      \rowcolor{improvebg} \quad \textit{vs. }Qwen3.5-27B & \textcolor{improvefg}{$\uparrow$4.08\%} & \textcolor{improvefg}{$\uparrow$3.77\%} & \textcolor{improvefg}{$\uparrow$15.34\%} \\
      \bottomrule
  \end{tabularx}
\end{table}

For the comparative analysis, we evaluate DisasterAgent against a wide range of mainstream general domain VLMs, specialized remote sensing VLMs, and existing remote sensing agent frameworks. To ensure a fair comparison and isolate the performance gains attributed to our proposed workflow, it is crucial to note that we equip the baseline remote sensing agents with the exact same specialized vision expert toolkit used by DisasterAgent. This rigorous setting guarantees that any observed improvements stem directly from our structured spatiotemporal memory and Evidence-based Augmentation mechanisms, rather than differences in the underlying perception tools. For the LLM as a judge evaluation, we utilize DeepSeek-V4-Flash \cite{deepseekai2026deepseekv4} to score the generated reports. The evaluation prompts strictly follow the settings of the DisasterM3 benchmark \cite{wang2026disasterm3}.

\subsection{Performance Comparison}

\begin{table*}[!t]
  \centering
  \caption{Performance Comparison on the DisasterM3 Benchmark}
  \label{tab:disasterm3}
  \setlength{\tabcolsep}{8pt}
  \renewcommand{\arraystretch}{1.3}
  \begin{tabular}{l|c|c|c|cccc|cccc}
      \toprule
      \multirow{3}{*}{Method} & \multicolumn{1}{c|}{\multirow{2}{*}{\textbf{DTR}}} & \multicolumn{1}{c|}{\multirow{2}{*}{\textbf{BDC}}} & \multicolumn{1}{c|}{\multirow{2}{*}{\textbf{DRE}}} & \multicolumn{8}{c}{\textbf{Disaster Report Generation}} \\
      \cmidrule(lr){5-12}
      & & & & \multicolumn{4}{c|}{\textbf{Disaster Captioning}} & \multicolumn{4}{c}{\textbf{Restoration Advice}} \\
      \cmidrule(lr){2-2} \cmidrule(lr){3-3} \cmidrule(lr){4-4} \cmidrule(lr){5-8} \cmidrule(lr){9-12}
      & ACC$\uparrow$ & MAE$\downarrow$ & MAE$\downarrow$ & AVG$\uparrow$ & DAP$\uparrow$ & DDR$\uparrow$ & FC$\uparrow$ & AVG$\uparrow$ & RN$\uparrow$ & APP$\uparrow$ & SC$\uparrow$ \\
      \midrule
      \multicolumn{12}{l}{\textcolor{blue}{$\bullet$} \textbf{General VLM}} \\
      Kimi K2.6 \cite{team2025kimi}        & 74.22 & 32.63 & 4.46 & 1.37 & 1.61 & 1.21 & 1.30 & 2.96 & 3.52 & 2.78 & 2.57 \\
      MiniMax-M3 \cite{lai2026minimax}     & 68.84 & 26.50 & \underline{3.81} & 2.03 & 2.31 & 1.80 & 1.97 & 2.79 & 3.37 & 2.58 & 2.42 \\
      Qwen3.5-27B \cite{qwen_scope}     & 70.25 & 30.05 & 4.81 & 1.88 & 2.16 & 1.65 & 1.81 & 2.86 & 3.47 & 2.67 & 2.44 \\
      Qwen3.5-9B \cite{qwen_scope}       & 64.02 & 35.66 & 14.38 & 1.71 & 2.02 & 1.47 & 1.65 & 2.62 & 3.23 & 2.43 & 2.19 \\
      InternVL3.5-8B \cite{wang2025internvl3}    & 47.03 & 36.70 & 26.16 & 1.16 & 1.43 & 0.97 & 1.09 & 2.55 & 3.26 & 2.26 & 2.14 \\
      \midrule
      \multicolumn{12}{l}{\textcolor[rgb]{0.8,0.5,0}{$\bullet$} \textbf{Remote Sensing VLM}} \\
      GeoChat-7B \cite{kuckreja2024geochat}        & 42.49 & 32.79 & 7.07 & 0.78 & 0.97 & 0.34 & 1.03 & 1.72 & 2.18 & 1.60 & 1.39 \\
      TeoChat-7B \cite{irvin2024teochat}          & 61.76 & 34.03 & 11.24 & 1.29 & 1.56 & 0.64 & 1.65 & 1.62 & 2.11 & 1.47 & 1.29 \\
      \midrule
      \multicolumn{12}{l}{\textcolor[rgb]{0.50,0.20,0.65}{$\bullet$} \textbf{Remote Sensing Agent} \textnormal{\textit{(with the same expert toolkit as ours)}}} \\
      EarthAgent \cite{feng2026earth}         & \underline{95.18} & \textbf{18.82} & \textbf{2.21} & \underline{2.36} & \underline{2.72} & \underline{2.03} & \underline{2.34} & \underline{3.15} & \underline{3.76} & \underline{2.96} & \underline{2.74} \\
      OpenEarthAgent \cite{shabbir2026openearthagent}        & 94.62 & \textbf{18.82} & \textbf{2.21} & 2.28 & 2.61 & 1.94 & 2.29 & 3.13 & 3.76 & 2.92 & 2.70 \\
      \midrule
      \multicolumn{12}{l}{\textcolor{green}{$\bullet$} \textbf{Ours}} \\
      DisasterAgent     & \textbf{95.47} & \underline{19.00} & \textbf{2.21} & \textbf{2.87} & \textbf{3.20} & \textbf{2.52} & \textbf{2.88} & \textbf{3.22} & \textbf{3.77} & \textbf{3.05} & \textbf{2.84} \\
      \rowcolor{improvebg} \quad \textit{vs. }Qwen3.5-27B & \textcolor{improvefg}{$\uparrow$25.22\%} & \textcolor{improvefg}{$\downarrow$11.05} & \textcolor{improvefg}{$\downarrow$2.60} & \textcolor{improvefg}{$\uparrow$19.8\%} & \textcolor{improvefg}{$\uparrow$20.8\%} & \textcolor{improvefg}{$\uparrow$17.4\%} & \textcolor{improvefg}{$\uparrow$21.4\%} & \textcolor{improvefg}{$\uparrow$7.2\%} & \textcolor{improvefg}{$\uparrow$6.0\%} & \textcolor{improvefg}{$\uparrow$7.6\%} & \textcolor{improvefg}{$\uparrow$8.0\%} \\
      \bottomrule
  \end{tabular}
\end{table*}

As shown in Table~\ref{tab:RSCC}, DisasterAgent consistently surpasses all zero-shot VLM baselines across all metrics on the RSCC benchmark. Compared to the Qwen3.5-27B base model, it achieves absolute improvements of 4.08\%, 3.77\%, and 15.34\% on ROUGE-L, METEOR, and ST5-SCS, respectively. Remarkably, on the contextual similarity metric, it significantly outperforms the fine-tuned RSCCM-7B model on ST5-SCS (67.18\% vs. 58.52\%), despite not utilizing any fine-tuning for specific tasks. Regarding the $n$-gram based metrics, DisasterAgent achieves a higher ROUGE-L score (15.76\% vs. 14.99\%) than RSCCM-7B, while RSCCM-7B retains a slight edge on METEOR (16.05\% vs. 14.77\%). This slight edge is expected given its training objective tailored for RSCC, which directly optimizes for exact token matching. Furthermore, when compared to other remote sensing agent frameworks (EarthAgent and OpenEarthAgent) equipped with the same expert toolkit, DisasterAgent achieves the highest ST5-SCS (67.18\%) and highly competitive ROUGE-L and METEOR scores. The substantial gain in ST5-SCS underscores that DisasterAgent produces semantically richer and more accurate change descriptions. This improvement is primarily attributed to the STPG and SCE modules, which explicitly capture and reason over spatiotemporal evolution rather than relying on fragmented tool outputs or end-to-end caption generation.

On the DisasterM3 benchmark, we evaluate DisasterAgent against a diverse set of baselines, including general VLMs, remote sensing VLMs, and remote sensing agents. As shown in Table~\ref{tab:disasterm3}, DisasterAgent achieves highly competitive performance across all task families, with particularly remarkable advantages in comprehensive report generation. 

\begin{figure}[!t]
  \centering
  \includegraphics[width=0.84\columnwidth]{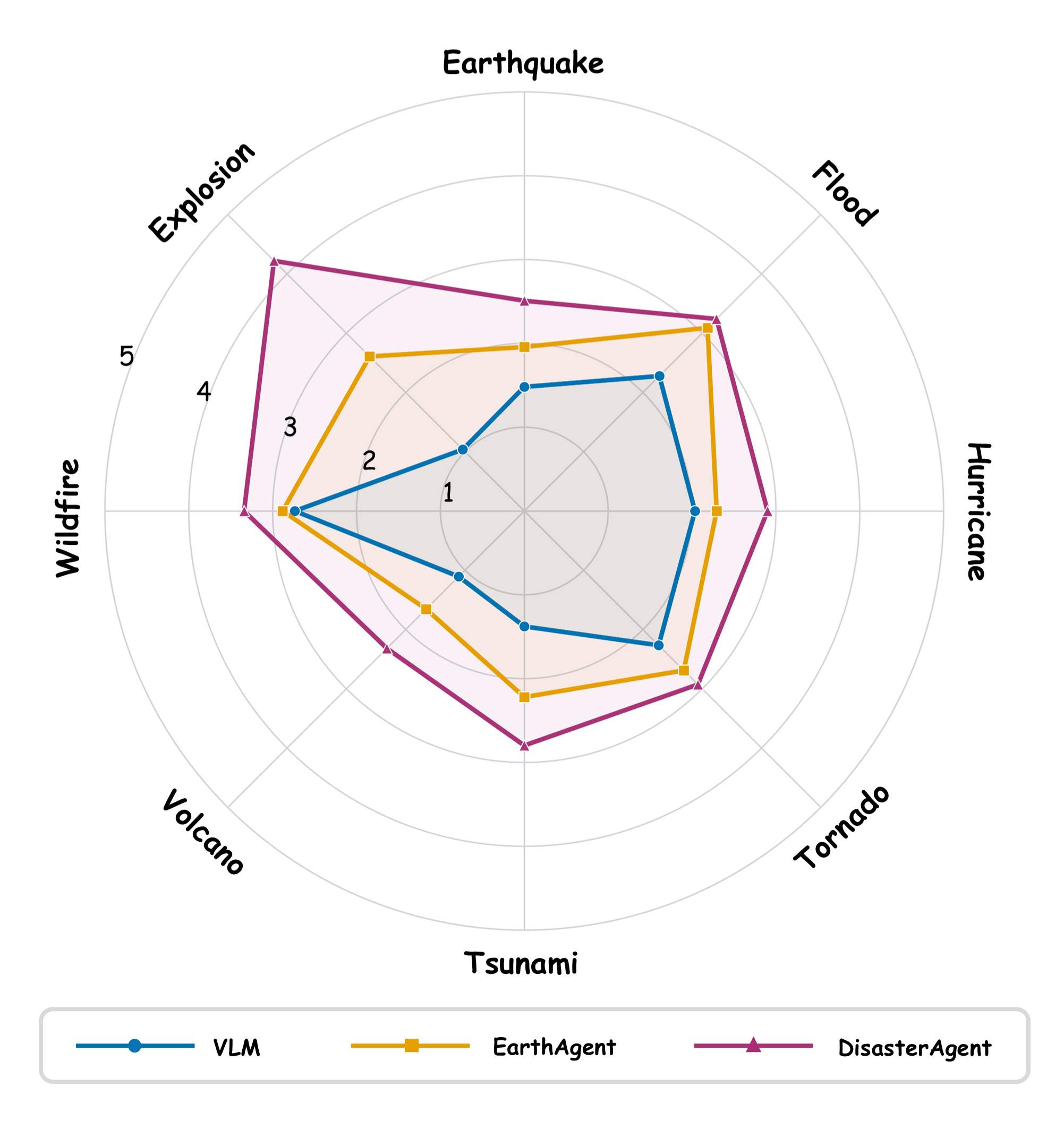}
  \caption{Disaster captioning scores on the DisasterM3 benchmark across eight disaster types. The VLM, EarthAgent, and DisasterAgent all use Qwen3.5-27B as the base VLM, and EarthAgent and DisasterAgent both use the same expert toolkit.}
  \label{fig:radar}
  \vspace{-8pt}
\end{figure}

Compared with the base model Qwen3.5-27B (which serves as the central brain in our framework), DisasterAgent improves the average score for disaster report generation by 13.5\%, with increases of 19.8\% for Disaster Captioning and 7.2\% for Restoration Advice. This substantial improvement highlights that the proposed evidence-driven workflow, integrating visual grounding, semantic verification, and historical experience, effectively mitigates the hallucination issues common in pure VLM generation. Furthermore, among the open source baselines, smaller models (e.g., InternVL3.5-8B, TeoChat-7B) lag considerably behind in generating coherent and factually correct reports. Similarly, while commercial APIs (e.g., Kimi K2.6, MiniMax-M3) demonstrate competitive reasoning capabilities, they still fall short of DisasterAgent's performance. This indicates that neither general domain pre-training nor massive model scale is sufficient for reliable disaster assessment without structured reasoning. Crucially, when comparing agentic frameworks, DisasterAgent, EarthAgent, and OpenEarthAgent achieve comparable performance in basic recognition and quantitative tasks (DTR, BDC, and DRE). This alignment is expected since all three agents are equipped with the identical vision expert toolkit, which provides the same foundational perception capabilities. However, when focusing on the highly complex comprehensive report generation tasks, DisasterAgent demonstrates a distinct advantage. It outperforms EarthAgent by absolute margins of 0.51 and 0.07 on the average scores for Disaster Captioning and Restoration Advice, respectively. This proves that simply stacking perception tools is inadequate; the structured STPG memory and multi-source evidence augmentation are vital for synthesizing fragmented tool outputs into high quality reports. Fig.~\ref{fig:radar} further compares disaster captioning scores across eight DisasterM3 types. DisasterAgent outperforms both Qwen3.5-27B and EarthAgent on every disaster type, with the largest margins on explosion, earthquake, and tsunami scenes.

\begin{table}[!t]
  \centering
  \caption{Ablation Study on the Structural Components}
  \label{tab:module_ablation}
  \renewcommand{\arraystretch}{1.2}
  \begin{tabular*}{\columnwidth}{@{\extracolsep{\fill}}lccccc}
    \toprule
    \textbf{Group} & \textbf{STPG} & \textbf{EA} & \textbf{\makecell{Disaster \\ Captioning}} & \textbf{\makecell{Restoration \\ Advice}} & \textbf{AVG} \\
    \midrule
    w/o Toolkit & $\times$ & $\times$ & 1.88 & 2.86 & 2.37 \\
    \midrule
    \multirow{3}{*}{w/ Toolkit} 
      & $\times$ & $\times$ & 2.36 & 3.15 & 2.76 \\
      & \checkmark & $\times$ & 2.59 & 2.96 & 2.78 \\
      & \checkmark & \checkmark & \textbf{2.87} & \textbf{3.22} & \textbf{3.05} \\
    \bottomrule
  \end{tabular*}
\end{table}

\subsection{Ablation Studies}

To systematically evaluate the contribution of each proposed module, we conduct comprehensive ablation studies on the disaster report generation tasks of DisasterM3-Bench. All ablation variants utilize the same Qwen3.5-27B base model, DeepSeek-V4-Flash judge, and decoding configurations, with only the specific framework components toggled. The ablation studies are divided into two parts: first, we investigate the effectiveness of the structural components; second, we conduct a detailed analysis of the individual evidence streams.

\begin{figure}[!t]
  \centering
  \includegraphics[width=\linewidth]{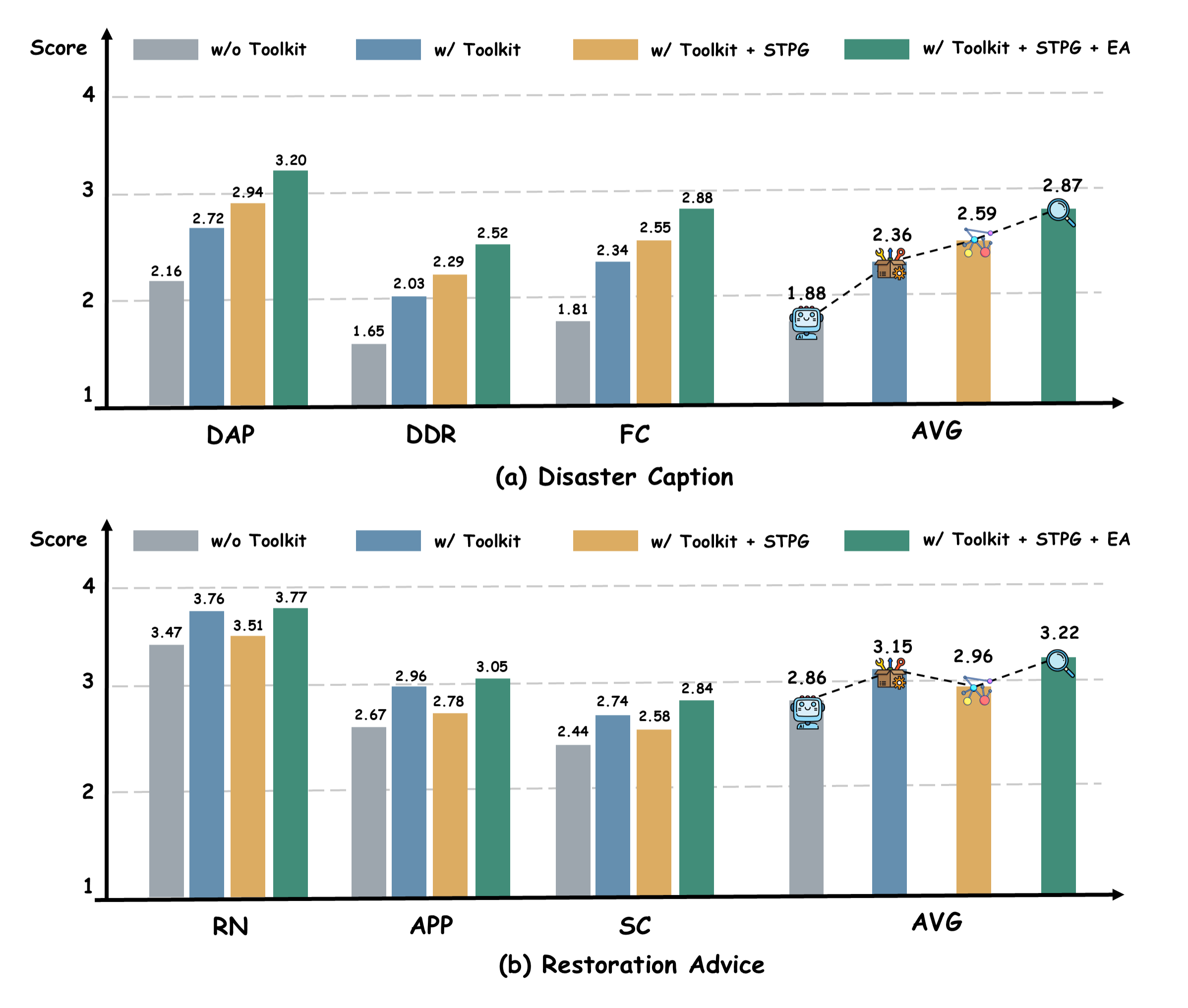}
  \caption{Detailed ablation results across individual evaluation metrics on the DisasterM3 benchmark. The figure illustrates the incremental performance improvements in (a) disaster captioning and (b) restoration advice when integrating the expert toolkit, STPG, and EA into the base model.}
  \label{fig:ablation}
  \vspace{-8pt}
\end{figure}

\subsubsection{Effectiveness of Structural Components}

As shown in Table~\ref{tab:module_ablation}, we first establish a \textit{w/o Toolkit} baseline, where the VLM generates reports end-to-end without any tool invocation, yielding an overall average score of 2.37. By equipping the VLM with a specialized vision expert toolkit via a standard Plan-and-Execute agent, the average score is raised to 2.76. To isolate the impact of these perception tools and specifically evaluate the performance gains attributed to our reasoning workflow, the remaining two settings form a controlled group that shares the exact same expert toolkit. Introducing the STPG to organize the tool outputs primarily boosts the disaster captioning performance, which drives the overall average up to 2.78. Finally, enabling the complete Evidence-based Augmentation (EA) produces the full model with a peak average score of 3.05.

\begin{figure*}[!t]
  \centering
  \includegraphics[width=\linewidth]{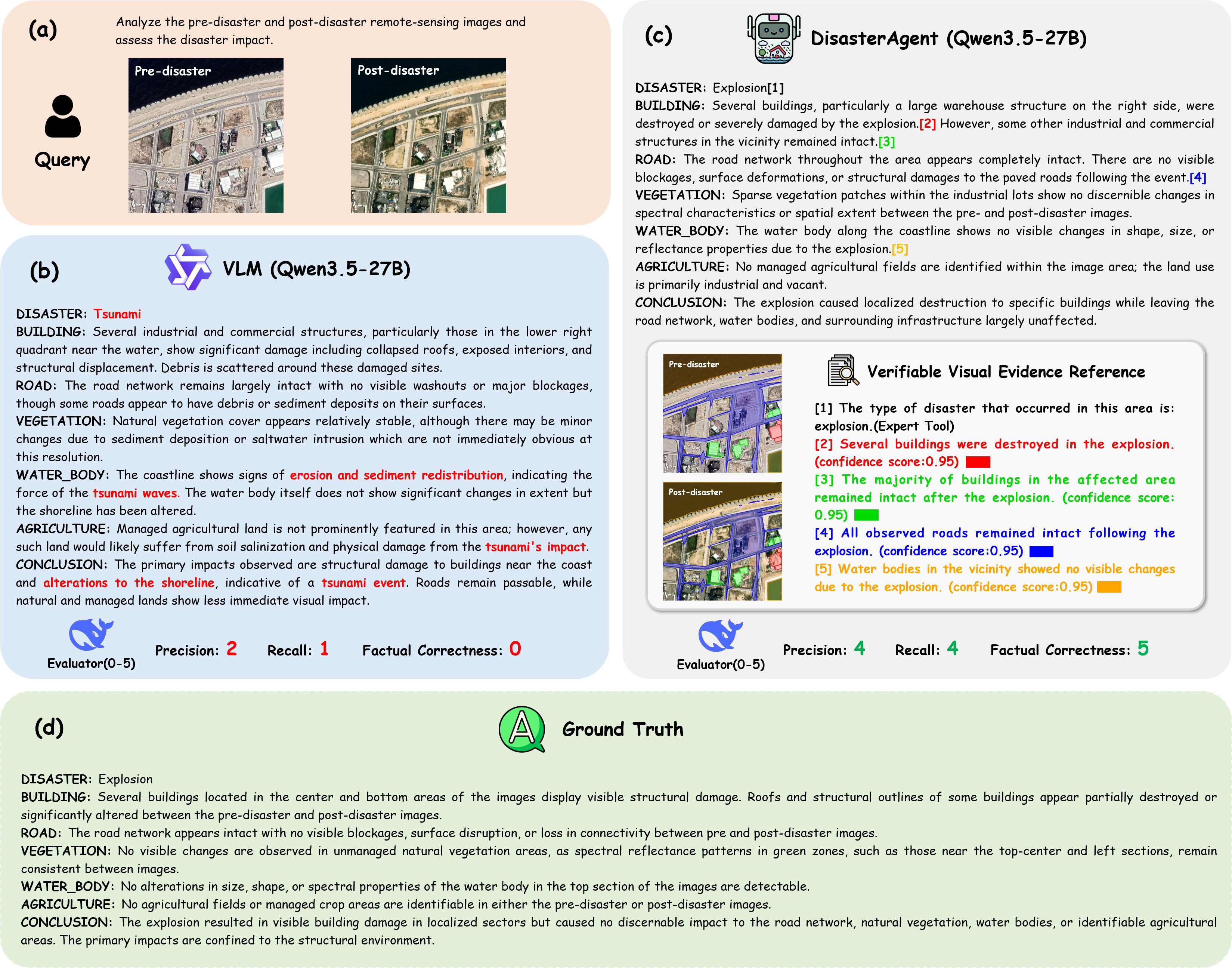}
  \caption{Qualitative results on a DisasterM3 report generation case. (a) The input query with the corresponding pre- and post-disaster images. (b) Caption produced by Qwen3.5-27B, which attributes the scene to a tsunami and introduces unsupported coastal changes (scores: DAP 2, DDR 1, FC 0). (c) Caption produced by DisasterAgent, which identifies a localized explosion, notes that surrounding roads and water bodies remain intact, and links each claim to a numbered visual evidence reference (scores: DAP 4, DDR 4, FC 5). (d) Ground truth caption.}
  \label{fig6_Qualitative}
  \vspace{-8pt}
\end{figure*}

To understand the specific impact of these components on detailed metrics, we analyze the performance trends illustrated in Fig.~\ref{fig:ablation}. The transition from the toolkit free baseline to the Plan-and-Execute agent confirms that specialized perception tools provide essential visual cues. However, without a unified spatiotemporal structure, the tool outputs remain fragmented, limiting the disaster captioning score to 2.36. As depicted in Fig.~\ref{fig:ablation}(a), introducing the STPG significantly improves disaster captioning to 2.59, demonstrating that explicitly encoding spatial relations and temporal evolution is highly beneficial for factual descriptions. Conversely, Fig.~\ref{fig:ablation}(b) shows that restoration advice drops slightly from 3.15 to 2.96 at this stage, suggesting that a structured perception brief without historical priors is less effective for action oriented recommendations. Finally, integrating EA (which comprises SCE, VGE, and EEE) maximizes performance across all dimensions, reaching 2.87 in captioning and 3.22 in restoration advice. Overall, these detailed trends prove that while the expert toolkit is necessary, structured spatiotemporal perception and evidence-driven generation are indispensable for converting fragmented tool outputs into highly verifiable disaster reports.

\subsubsection{Contribution of Multi-Source Evidence}

To understand the specific role of each evidence stream from the perspective of final report generation, we perform an ablation study on the full model by removing one component at a time, as detailed in Table~\ref{tab:evidence_ablation_detail}. Removing EEE results in the most noticeable performance degradation, with the Disaster Captioning average score dropping to 2.56 and the Restoration Advice average score decreasing to 3.08. This confirms that EEE serves as a crucial reference, providing essential domain-specific guidance for formulating normative and actionable restoration advice. Furthermore, the substantial drop in the Disaster Captioning average score reveals that historical context is also vital for guiding the VLM to generate accurate factual descriptions of disaster scenarios.

Removing SCE causes the Disaster Captioning average score to fall to 2.58. Notably, this removal leads to a substantial decline in DAP and FC. This validates that the confidence-scored, self-auditing SCE mechanism serves as the primary verifiability anchor, effectively filtering out hallucinations during the caption generation stage.

Removing VGE has a relatively marginal effect on the overall Disaster Captioning average score. However, it causes a noticeable drop in DDR. This indicates that while SCE and EEE provide strong semantic and historical priors, VGE supplies fine-grained, pixel-level localization that is essential for accurately grounding damage quantities and spatial distribution.

\begin{table}[!t]
  \centering
  \caption{Contribution of Multi-Source Evidence}
  \label{tab:evidence_ablation_detail}
  \renewcommand{\arraystretch}{1.3}
  \begin{tabular*}{\columnwidth}{@{\extracolsep{\fill}}clccccc}
    \toprule
    \multirow{2}{*}{\textbf{Task}} & \multirow{2}{*}{\textbf{Metric}} & \multirow{2}{*}{\textbf{Full}} & \multicolumn{3}{c}{\textbf{w/o (Remove)}} & \multirow{2}{*}{\textbf{Base}} \\
    \cmidrule(lr){4-6}
    & & & EEE & VGE & SCE & \\
    \midrule
    \multirow{4}{*}{\rotatebox{90}{\shortstack{Disaster\\Captioning}}} 
    & AVG & \textbf{2.87} & 2.56 & 2.85 & 2.58 & 1.88 \\
    & DAP & 3.20 & 2.94 & \textbf{3.22} & 2.87 & 2.16 \\
    & DDR & \textbf{2.52} & 2.14 & 2.46 & 2.30 & 1.65 \\
    & FC  & 2.88 & 2.60 & \textbf{2.89} & 2.56 & 1.81 \\
    \midrule
    \multirow{4}{*}{\rotatebox{90}{\shortstack{Restoration\\Advice}}} 
    & AVG & \textbf{3.22} & 3.08 & - & - & 2.86 \\
    & RN  & \textbf{3.77} & 3.68 & - & - & 3.47 \\
    & APP & \textbf{3.05} & 2.90 & - & - & 2.67 \\
    & SC  & \textbf{2.84} & 2.65 & - & - & 2.44 \\
    \bottomrule
  \end{tabular*}
\end{table}

\subsection{Qualitative Analysis}
To intuitively demonstrate the advantages of our proposed framework, we present a qualitative comparison of disaster captioning between the base VLM (Qwen3.5-27B) and DisasterAgent, as illustrated in Fig.~\ref{fig6_Qualitative}. 

When relying on direct end-to-end inference, the base VLM suffers from severe hallucinations due to its black-box nature and lack of explicit spatiotemporal grounding. In this specific case, the base model misclassifies the event as a ``Tsunami'' rather than an explosion. Consequently, it fabricates nonexistent details that align with its erroneous assumption, such as ``erosion and sediment redistribution'' along the coastline and ``alterations to the shoreline.'' This unconstrained generation leads to extremely poor evaluation scores, including a Factual Correctness of 0, highlighting the unreliability of standard VLMs for rigorous remote sensing tasks.

In contrast, DisasterAgent leverages its evidence-driven architecture to produce a highly accurate and verifiable disaster report. It correctly identifies the disaster as an ``Explosion'' and precisely localizes the damage, noting that a ``large warehouse structure on the right side'' was destroyed while the surrounding roads and water bodies remained intact. Beyond mere accuracy, the most significant advantage of DisasterAgent is its verifiability. As shown in the generated report, every critical statement is appended with a citation tag. These tags correspond to a structured ``Verifiable Visual Evidence Reference'' that links directly back to the explicitly rendered perception masks and confidence scores from the STPG. By grounding its generation in traceable visual and semantic evidence, DisasterAgent effectively mitigates hallucinations, achieving a perfect Factual Correctness score of 5 and providing the verifiability required for real-world emergency response. To complement the qualitative analysis, we also provide four additional disaster captioning examples covering tsunami, earthquake, flood, and wildfire scenarios in Appendix~\ref{app:qual}, as shown in Figs.~\ref{fig:app_tsunami}--\ref{fig:app_wildfire}.

\begin{table}[!t]
  \centering
  \caption{SCE Citation Coverage and Entity Associations}
  \label{tab:evidence_statistics}
  \renewcommand{\arraystretch}{1.2}
  \begin{tabular*}{\columnwidth}{@{\extracolsep{\fill}}lcc}
    \toprule
    \textbf{Metric} & \textbf{DisasterM3} & \textbf{RSCC} \\
    \midrule
    Number of Samples & 2,004 & 988 \\
    Samples with SCE Citations (\%) & 94.86 & 85.32 \\
    Mean STPG Entities per Sample  & 182.37 & 35.84 \\
    Mean SCE Citations per Sample & 3.40 & 1.93 \\
    Mean Linked STPG Entities per SCE Citation & 5.23 & 8.02 \\
    \bottomrule
  \end{tabular*}
\end{table}

\subsection{Evidence Verifiability Analysis}

\begin{figure}[!t]
  \centering
  \includegraphics[width=\columnwidth]{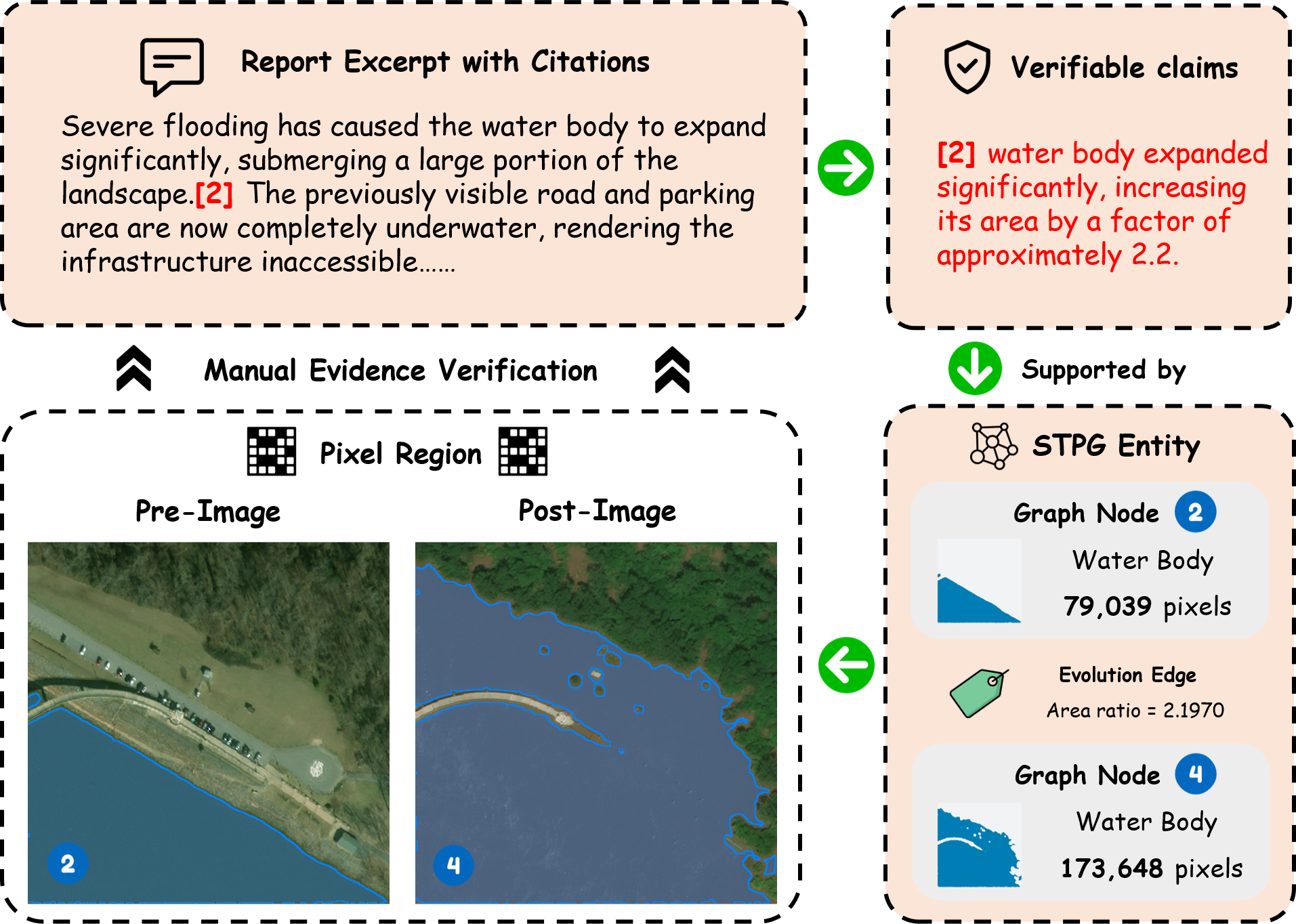}
  \caption{An SCE citation and its linked STPG entities. The citation in the report excerpt is traced to the corresponding SCE claim and its associated STPG entities, including region nodes and an evolution edge, which are rendered on the pre- and post-disaster images.}
  \label{fig7_evidence}
  \vspace{-8pt}
\end{figure}

As reported in Table~\ref{tab:evidence_statistics}, SCE citations cover most evaluated samples on both benchmarks, and each citation is associated with multiple STPG entities. Fig.~\ref{fig7_evidence} shows how an SCE citation is traced to its linked STPG entities, using a flood example from the RSCC benchmark. In the report excerpt, the citation connects the description of water-body expansion to the corresponding SCE claim. This claim is associated with STPG entities comprising pre-disaster region node 2, post-disaster region node 4, and their evolution edge, which are rendered on the pre- and post-disaster images. The water-body masks contain 79,039 and 173,648 pixels, respectively, giving a post-to-pre area ratio of approximately 2.197. This ratio is consistent with the SCE claim that the water-body area expanded to approximately 2.2 times its pre-disaster extent, and the rendered masks make the locations and boundaries of these regions directly inspectable. The example shows how an SCE citation is linked to its STPG entities and the corresponding image regions, providing explicit spatiotemporal provenance for verification.

\section{Conclusion}

In this paper, we present DisasterAgent, an evidence-driven agent that elevates remote sensing disaster assessment from basic visual perception to verifiable comprehensive report generation. The framework couples a specialized vision expert toolkit with the STPG, which systematically organizes fragmented tool outputs into coherent spatial relationships and temporal evolution edges. A multi-source evidence augmentation mechanism further integrates visual grounding, semantic verification, and historical experience to guide the generation process. Experiments on the DisasterM3 and RSCC benchmarks suggest the effectiveness of the proposed framework, particularly in mitigating hallucinations and anchoring factual claims to traceable evidence. In future work, we will extend DisasterAgent to multi modal data sources (e.g., Synthetic Aperture Radar) and multi temporal disaster imagery, enabling more resilient and dynamic disaster assessment systems.

\section*{Acknowledgments}
This work was supported by the National Natural Science Foundation of China under Grant 62425115, and partially supported by China Science and Technology Cloud (CSTCloud).

\bibliographystyle{IEEEtran}
\bibliography{references}

\clearpage
\appendices
\makeatletter
\twocolumn[{%
\section{Perception Tool Design}
\label{app:tools}
\vspace{1em}
\centering
\includegraphics[width=0.92\textwidth]{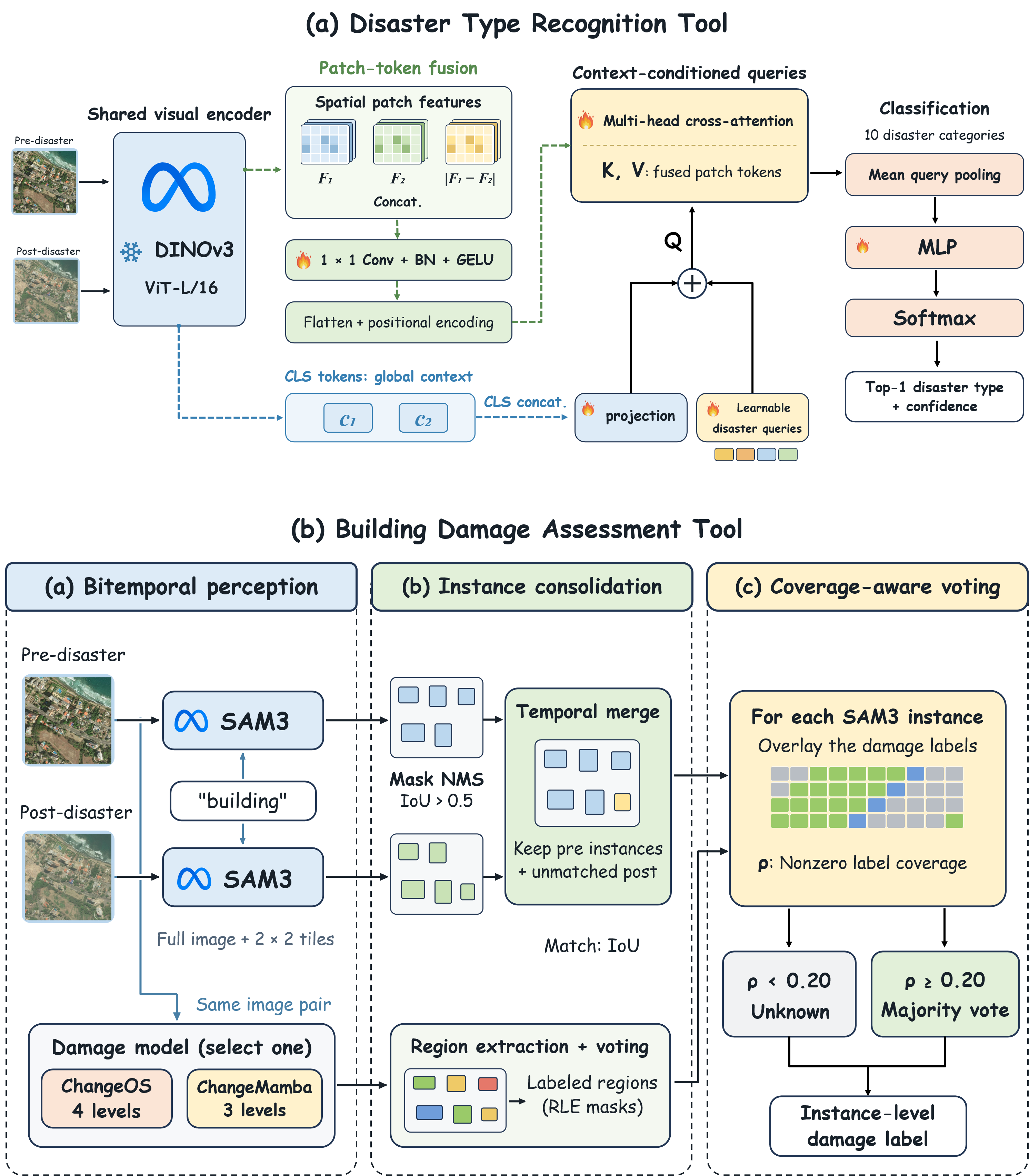}

\def\@captype{figure}%
\caption{Design of the perception tools. (a) Disaster type recognition encodes the pre- and post-disaster images with a shared DINOv3 ViT-L/16, fuses their patch tokens, and predicts one of ten disaster categories together with a confidence score. (b) Building damage assessment segments building instances with SAM3, consolidates them across the two dates, and assigns an instance-level damage label by coverage-aware voting over the masks from ChangeOS or ChangeMamba.}
\label{fig:app_tools}
\vspace{0.5\baselineskip}%
}]
\makeatother

\clearpage
\makeatletter
\twocolumn[{%
\section{Additional Qualitative Results}
\label{app:qual}
\parindent=1.0em
\parskip=0ex plus 0.2ex minus 0.1ex
\everypar{}%
To complement the qualitative analysis in the main text, we present four additional disaster captioning examples covering tsunami, earthquake, flood, and wildfire scenarios (Fig.~\ref{fig:app_tsunami}--\ref{fig:app_wildfire}).
\par

\vspace{1em}
\centering
\includegraphics[width=\textwidth]{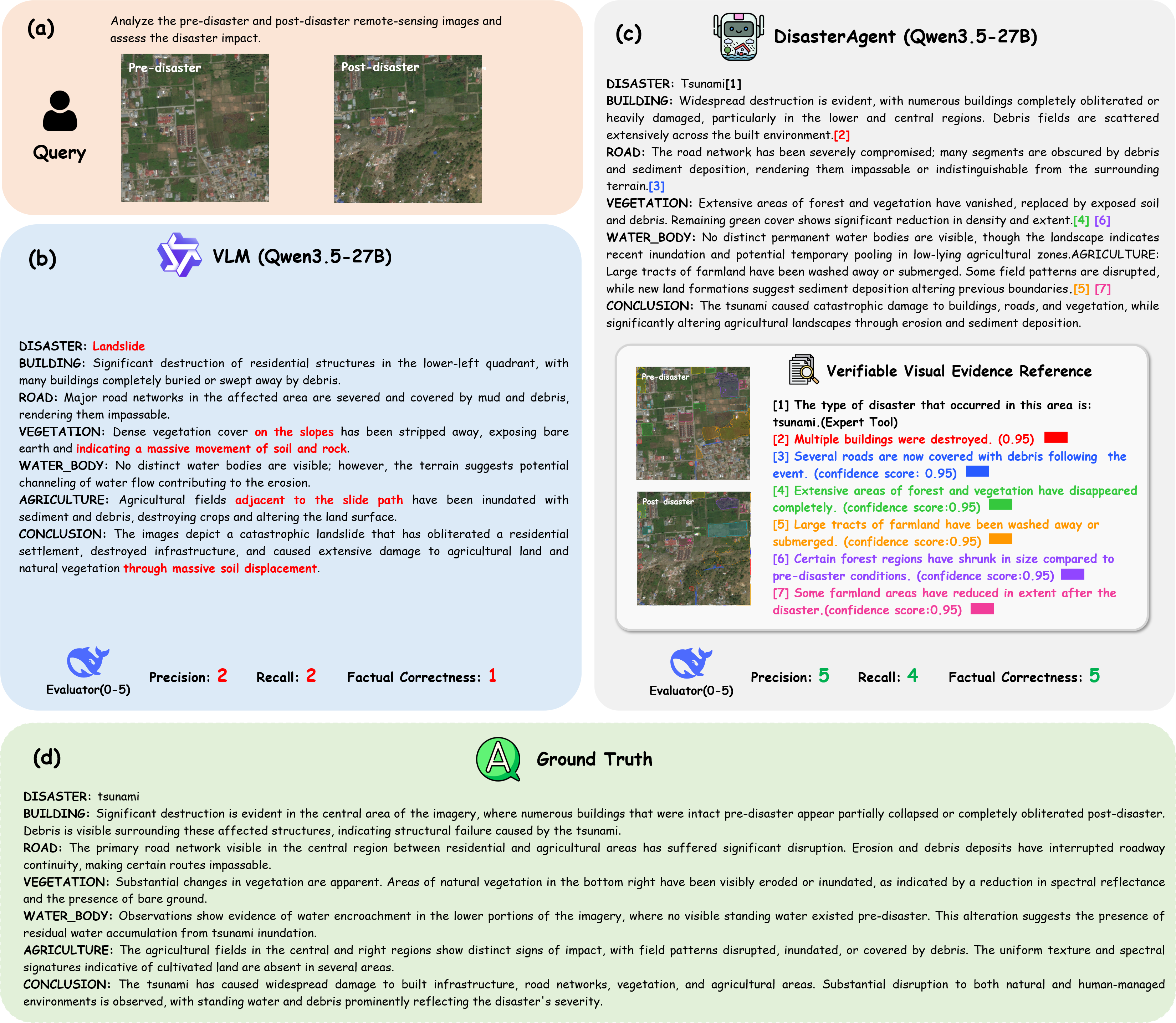}

\def\@captype{figure}%
\caption{Qualitative DisasterM3 captioning case for a tsunami scene. (a) The input query with the corresponding pre- and post-disaster images. (b) Caption produced by Qwen3.5-27B, which misclassifies the event as a landslide and introduces unsupported descriptions of slope-related soil and rock movement (scores: DAP 2, DDR 2, FC 1). (c) Caption produced by DisasterAgent, which identifies the tsunami and describes damage to buildings, roads, vegetation, and farmland, with selected claims linked to numbered evidence references (scores: DAP 5, DDR 4, FC 5). (d) Ground truth caption.}
\label{fig:app_tsunami}
\vspace{0.5\baselineskip}%
}]
\makeatother
\setlength{\columnwidth}{\dimexpr(\textwidth-\columnsep)/2\relax}
\setlength{\hsize}{\columnwidth}
\setlength{\linewidth}{\columnwidth}

\begin{figure*}[!t]
  \centering
  \includegraphics[width=\textwidth]{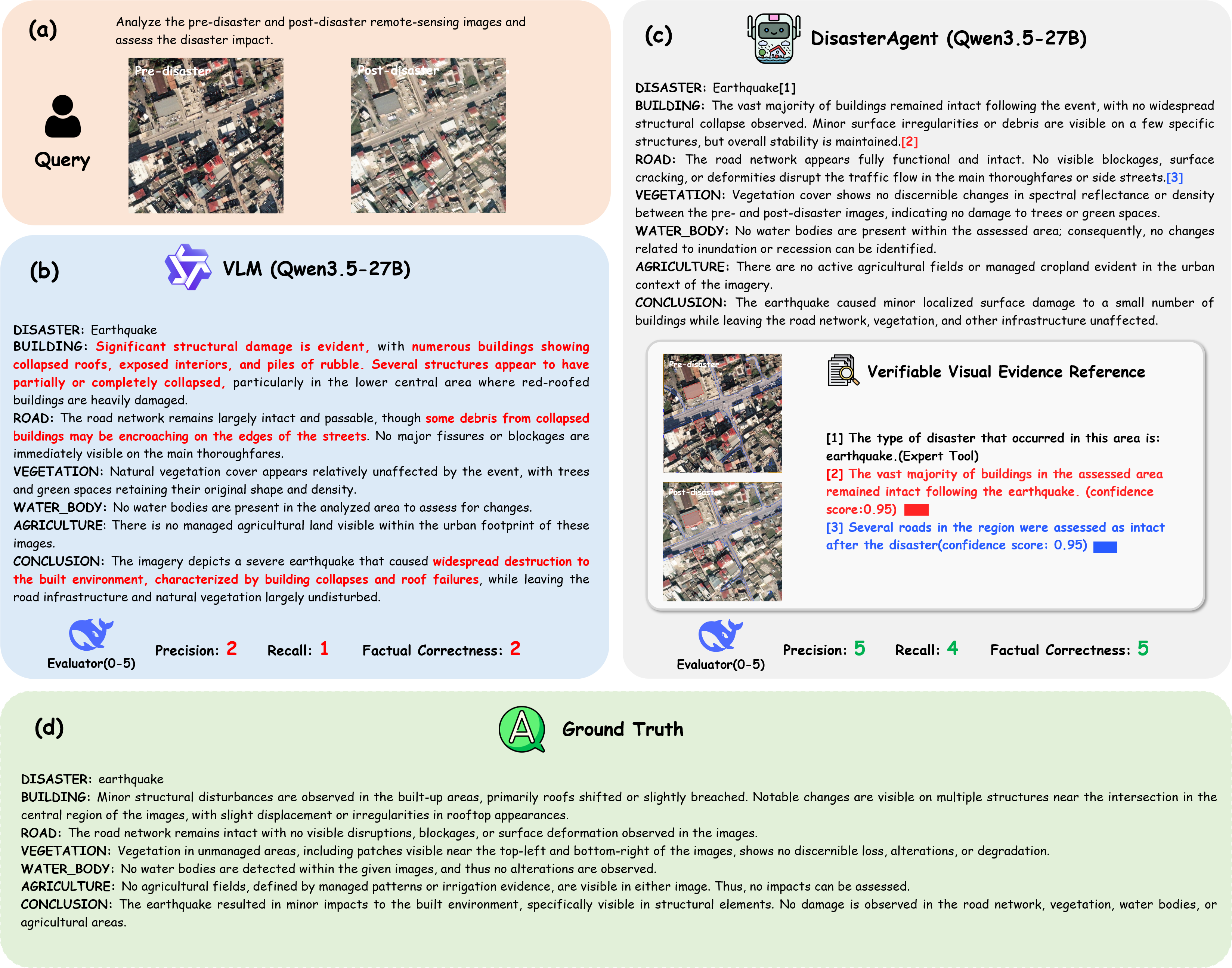}
  \caption{Qualitative DisasterM3 captioning case for an earthquake scene. (a) The input query with the corresponding pre- and post-disaster images. (b) Caption produced by Qwen3.5-27B, which exaggerates minor structural disturbances as widespread building collapse and severe destruction (scores: DAP 2, DDR 1, FC 2). (c) Caption produced by DisasterAgent, which describes buildings as largely intact, acknowledges limited structural disturbances, and identifies intact roads, with selected claims linked to numbered evidence references (scores: DAP 5, DDR 4, FC 5). (d) Ground truth caption.}
  \label{fig:app_earthquake}
\end{figure*}

\begin{figure*}[!t]
  \centering
  \includegraphics[width=\textwidth]{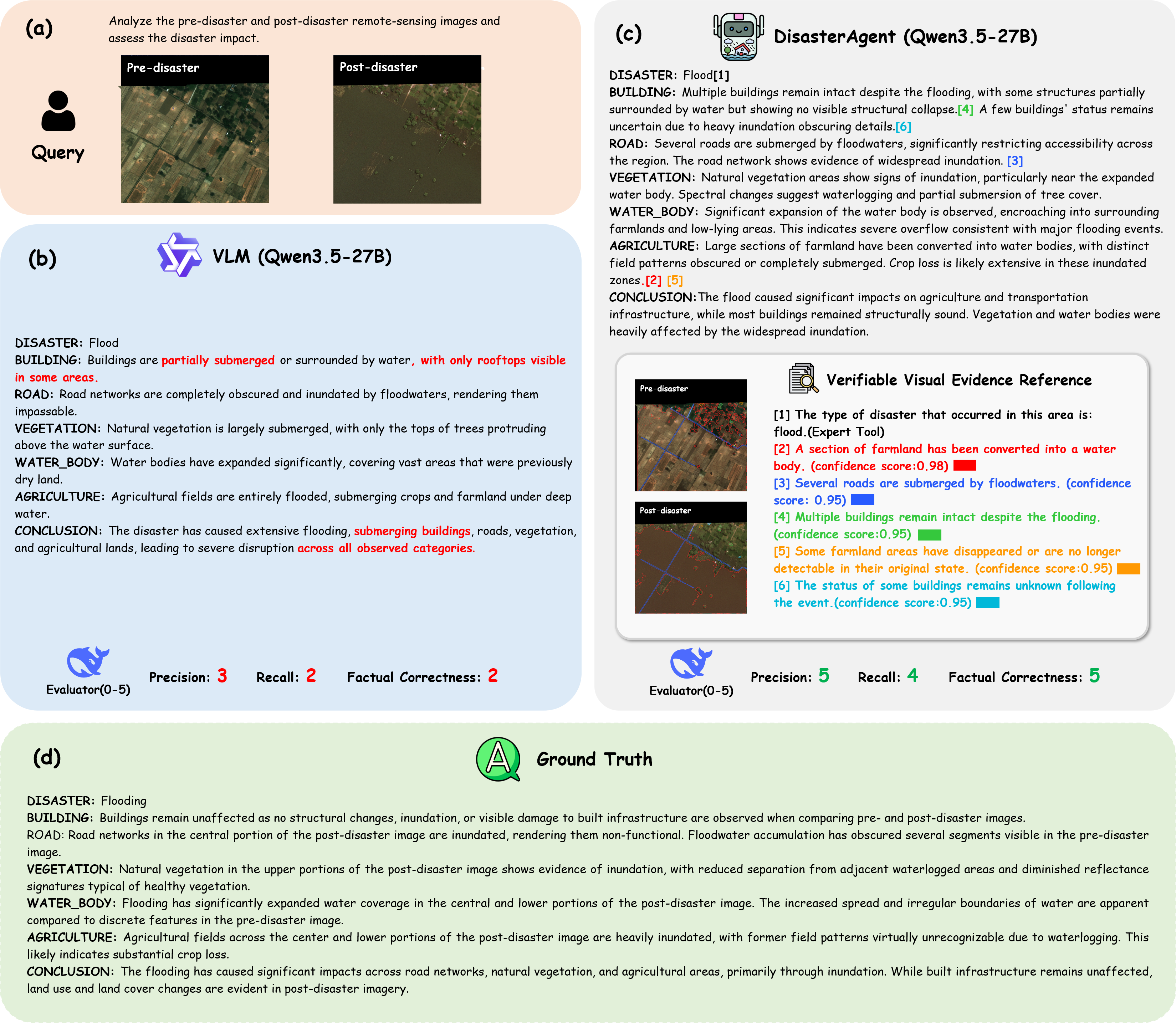}
  \caption{Qualitative DisasterM3 captioning case for a flood scene. (a) The input query with the corresponding pre- and post-disaster images. (b) Caption produced by Qwen3.5-27B, which identifies the flood but overstates its impact on buildings, including unsupported claims that some buildings are submerged with only their rooftops visible (scores: DAP 3, DDR 2, FC 2). (c) Caption produced by DisasterAgent, which distinguishes inundated roads and farmland from buildings assessed as intact and retains uncertainty for some structures, with selected claims linked to numbered evidence references (scores: DAP 5, DDR 4, FC 5). (d) Ground truth caption.}
  \label{fig:app_flood}
\end{figure*}

\begin{figure*}[!t]
  \centering
  \includegraphics[width=\textwidth]{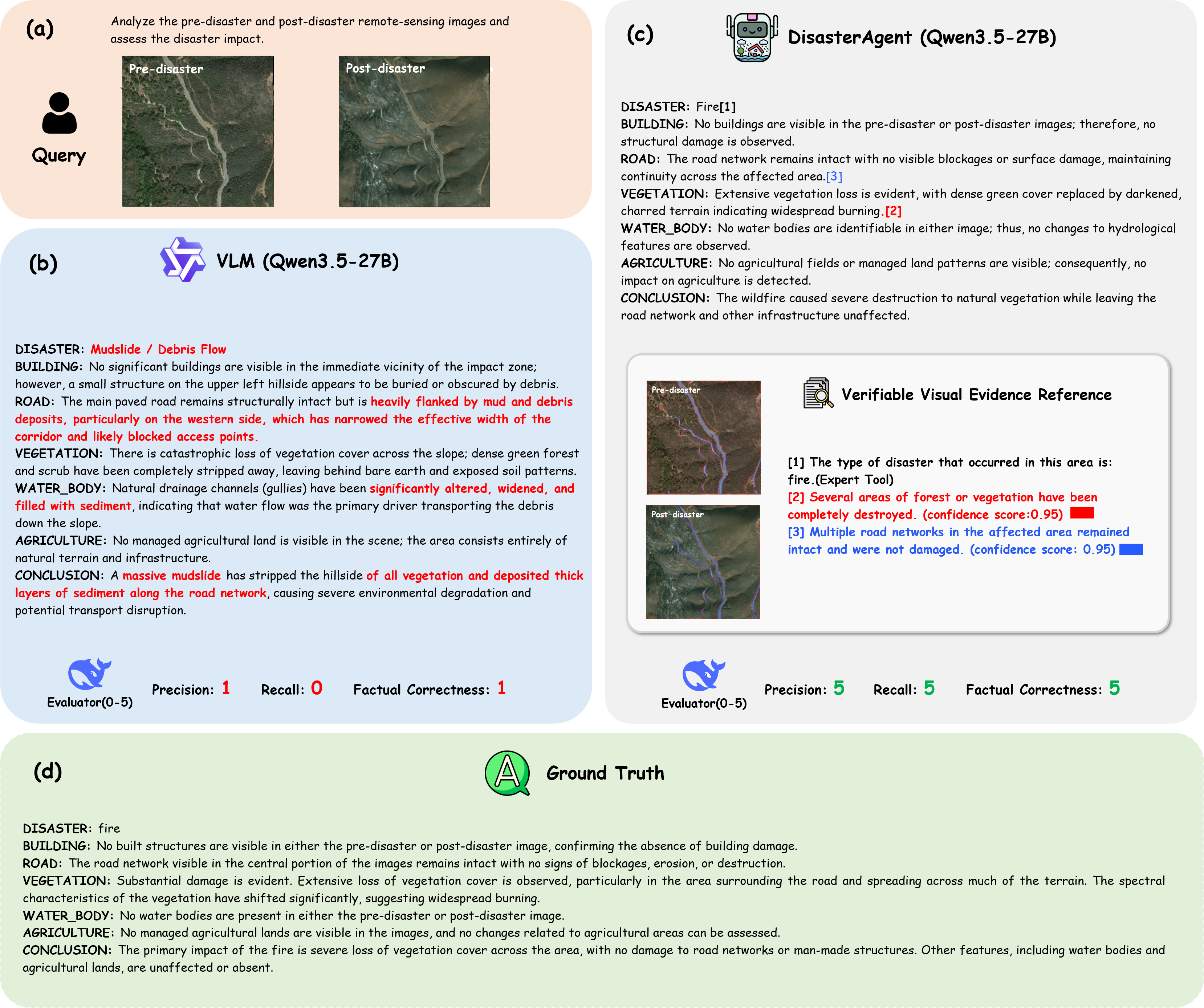}
  \caption{Qualitative DisasterM3 captioning case for a wildfire scene. (a) The input query with the corresponding pre- and post-disaster images. (b) Caption produced by Qwen3.5-27B, which misclassifies the event as a mudslide or debris flow and introduces unsupported claims of sediment deposition, altered drainage channels, and restricted road access (scores: DAP 1, DDR 0, FC 1). (c) Caption produced by DisasterAgent, which identifies the fire, describes extensive vegetation loss, and recognizes that the road network remains intact, with selected claims linked to numbered evidence references (scores: DAP 5, DDR 5, FC 5). (d) Ground truth caption.}
  \label{fig:app_wildfire}
\end{figure*}

\end{document}